%% file: sparse-transfer.tex
\documentclass[10pt,twocolumn,letterpaper]{article}

\usepackage[pagenumbers]{cvpr}
\makeatletter
\@namedef{ver@everyshi.sty}{}
\makeatother
\usepackage{tikz}

\input{math_commands.tex}

\usepackage{url}
\usepackage{todonotes}
\usepackage{amssymb}
\usepackage{booktabs}

\usepackage[pagebackref,breaklinks,colorlinks]{hyperref}

\renewcommand{\paragraph}[1]{\vspace{-0.1em} \smallskip\noindent {\bfseries\boldmath\ignorespaces #1}\hskip 0.9em plus 0.3em minus 0.3em}

\title{How Well Do Sparse ImageNet Models Transfer?}

\author{
Eugenia Iofinova\thanks{These authors contributed equally. \hspace{-1.7cm} Correspondence to: \quad \{\texttt{eugenia.iofinova, alexandra.peste}\}\texttt{@ist.ac.at}}  \\{\small IST Austria} \\ 
\and Alexandra Peste\footnotemark[1] \\ {\small IST Austria} \\ 
\and Mark Kurtz\\ {\small Neural Magic}\\
\and
Dan Alistarh\\
{\small IST Austria \&  Neural Magic} \\
}

\begin{document}

\maketitle
\begin{abstract}

Transfer learning is a classic paradigm by which models pretrained on large ``upstream'' datasets are adapted to yield good results on ``downstream'' specialized datasets. Generally,  more accurate  models on the ``upstream'' dataset tend to provide better transfer accuracy ``downstream''. 
In this work, we perform an in-depth investigation of this phenomenon in the context of convolutional neural networks (CNNs) trained on the ImageNet dataset, which have been pruned---that is, compressed by sparsifiying their connections. 
We consider transfer using unstructured pruned models obtained by applying several state-of-the-art pruning methods, including magnitude-based, second-order, re-growth, lottery-ticket, and regularization approaches, in the context of twelve standard transfer tasks. 
In a nutshell, our study shows that sparse models can match or even outperform the transfer performance of dense models, even at high sparsities, and, while doing so, can lead to significant inference and even training speedups. 
At the same time, we observe and analyze significant differences in the behaviour of different pruning methods.

\end{abstract}

\vspace{-1em}
\section{Introduction}

The large computational costs of deep learning have led to significant academic and industrial interest in \emph{model compression}, defined roughly as obtaining smaller-footprint models matching the accuracy of larger models. 
Model compression is a rapidly-developing area, and several general approaches have been investigated, of which pruning and quantization are among the most popular~\cite{hoefler2021sparsity, gholami2021survey}.

We focus our present study on \emph{weight pruning}, whose objective is to remove, by setting to zero, as many weights as possible without losing model accuracy.
Weight pruning is, arguably, the compression method with the richest history~\cite{lecun1990optimal} and is currently a very active research topic~\cite{hoefler2021sparsity}. 
Thanks to this trend, a set of fairly consistent accuracy benchmarks has emerged for pruning, along with increasingly efficient computational support \cite{mishra2021accelerating, graphcore, pmlr-v119-kurtz20a, elsen2020fast}. 

One major goal of model compression is to enable deployment on edge devices. Such devices may naturally encounter different data distributions, so it is tempting to ask how compressed models would perform for \emph{transfer learning}, broadly defined as leveraging information from some baseline ``upstream'' (``pretrained'') task in order to perform better on a ``downstream'' (``finetuning'') task. Specifically, we mainly focus on a prototypical transfer learning setup~\cite{kornblith2019better}: starting from models trained and compressed on the ImageNet-1K dataset~\cite{ImageNet}, we refine the resulting models onto several different target tasks. In this context, we examine the question of how well the resulting sparse models transfer. 
Our motivation is both practical---sparse transfer can provide speedups for both inference and training on the downstream model---and analytical, as we aim to shed light on the impact of sparsity on the resulting features. 

Our study will consider two common transfer learning variants: \emph{full finetuning}, where all unpruned weights can be optimized during transfer, 
and \emph{linear finetuning}, where only the final linear layer of the model is finetuned downstream. 
While both are popular, we will see that they can lead to different results. We additionally explore \emph{inference-time} speedups using a sparsity-aware inference engine~\cite{NM}, and for the first time examine \emph{training-time speed-up} achievable for linear finetuning via sparse models. Furthermore, we analyze the impact of different pruning methods and task characteristics on transfer performance.

We consider the top-performing pruning methods in terms of ImageNet accuracy, roughly split into three categories. 
The first is given by \emph{progressive sparsification} methods, which start from an accurate \emph{dense} baseline and proceed to gradually remove weights, followed by finetuning. 
The prototypical example is \emph{gradual magnitude pruning (GMP)}~\cite{hagiwara1994, han2015learning, zhu2017prune, gale2019state}, which uses absolute weight magnitude as the pruning criterion. 
In addition, we examine WoodFisher pruning~\cite{singh2020woodfisher}, which leverages second-order information for highly-accurate pruning. 

The second rough category is given by \emph{sparse regularized training} methods, which perform network compression, and possibly network re-growth, during the training process itself. 
The top-performing methods we consider here are Soft Threshold Reparametrization (STR)~\cite{kusupati2020soft}, Alternating Compressed/DeCompressed Training (AC/DC)~\cite{peste2021ac} and ``The Rigged Lottery'' (RigL)~\cite{evci2020rigging}.

The final category comprises Lottery Ticket Hypothesis (LTH)-style methods~\cite{frankle2018lottery, frankle2019stabilizing, chen2020lottery, chen2021lottery}. 
These methods emphasize the \emph{discovery} of sparse sub-networks, which can yield good accuracy when re-trained from scratch. 
Specifically, we consider LTH for transfer (LTH-T)~\cite{chen2021lottery}, which provides state-of-the-art results among such methods. 

We measure the transfer accuracy of sparse ImageNet models obtained via these pruning methods. Our main target application is given by twelve classic transfer datasets, described in Table~\ref{table:datasets}, ranging from general datasets, to more specialized ones. 
We mainly focus on the classic ResNet50~\cite{he2016deep} model, but we extend our analysis to ResNet18, ResNet34 and MobileNet-V1~\cite{howard2017mobilenets}, and we also  examine transfer performance for object detection tasks.

\paragraph{Contribution.} 
We present the first systematic study of how different pruning and transfer approaches impact transfer performance. 
\emph{Our main finding is that sparse models can consistently match the accuracy of the corresponding dense models on transfer tasks.} 
However, this behaviour is impacted by the following factors: \emph{pruning method} (e.g. regularization vs. progressive pruning), \emph{transfer approach} (full vs. linear), \emph{model sparsity} (e.g. moderate 80\% vs. high 98\% sparsity), and \emph{task type} (e.g. degree of specialization).

 \begin{table*}[h]
    \centering
    \scalebox{0.65}{
    \begin{tabular}{@{}cccccccccccccc@{}}
    \toprule
    Finetuning & Sparsity & Aircraft & Birds & Caltech-101 & Caltech-256 & Cars & CIFAR-10 & CIFAR-100 & DTD & Flowers & Food-101 & Pets & SUN397 \\
    \toprule
    Linear & 0\% & 49.2 $\pm$ 0.1 & 57.7$\pm$ 0.1 & 91.9$\pm$ 0.1 & \textbf{84.8$\pm$ 0.1} & 53.4$\pm$ 0.1 & 91.2$\pm$ 0. & \textbf{74.6$\pm$ 0.1}  & 73.5$\pm$ 0.2 & 91.6$\pm$ 0.1 & 73.2$\pm$ 0. & \textbf{92.6$\pm$ 0.1} & 60.1$\pm$ 0. \\
    
    & 80\% &
    55.2$\pm$ 0.2 & 
    58.4$\pm$ 0. & 
    \textbf{92.4$\pm$ 0.2} &
    \textbf{84.6$\pm$ 0.1} &
    58.6$\pm$ 0.1 &
    \textbf{91.4$\pm$ 0.} &
    \textbf{74.7$\pm$ 0.1} &
    \textbf{74.4$\pm$ 0.1} &
    \textbf{93.0$\pm$ 0.} &
    \textbf{73.9$\pm$ 0.} &
    \textbf{92.5 $\pm$ 0.1} &
    \textbf{60.4$\pm$ 0.} \\
    
    & 90\% &
    \textbf{56.6$\pm$ 0.1} &
    \textbf{58.7 $\pm$ 0.} &
    \textbf{92.5$\pm$ 0.1} &
    84.5$\pm$ 0.1 &
    \textbf{60.5 $\pm$ 0.1} & 
    91.0 $\pm$ 0. &
    74.3$\pm$ 0. &
    73.8$\pm$ 0.1 &
    \textbf{93.0 $\pm$ 0.1} &
    \textbf{73.8 $\pm$ 0.} &
    92.0 $\pm$ 0.1 &
    59.8 $\pm$ 0.1 \\
    \midrule
    
    Full &
    0\% & 83.6 $\pm$ 0.4 &
    72.4 $\pm$ 0.3 &
    \textbf{93.5 $\pm$ 0.1} &
    \textbf{86.1 $\pm$ 0.1} &
    \textbf{ 90.3 $\pm$ 0.2} &
    \textbf{97.4 $\pm$ 0.} &
    \textbf{85.6 $\pm$ 0.2}  &
    \textbf{76.2 $\pm$ 0.3} &
    95.0 $\pm$ 0.1 &
    \textbf{87.3 $\pm$ 0.1} &
    \textbf{93.4 $\pm$ 0.1} &
    \textbf{64.8 $\pm$ 0.}\\
    
    & 80\% &
    \textbf{84.8 $\pm$ 0.2} &
    \textbf{73.4 $\pm$ 0.1} &
    \textbf{93.7 $\pm$ 0.1} &
    85.4 $\pm$ 0.2 &
    \textbf{90.5 $\pm$ 0.2} &
    97.2 $\pm$ 0.1 &
    85.1 $\pm$ 0.1 &
    \textbf{75.7 $\pm$ 0.5} &
    \textbf{96.1 $\pm$ 0.1} &
    \textbf{87.4 $\pm$ 0.1} &
    \textbf{93.4 $\pm$ 0.1} &
    64.0 $\pm$ 0. \\ 
    
    & 90\% &
    \textbf{84.9 $\pm$ 0.3 } &
    72.9 $\pm$ 0.2 &
    \textbf{93.9 $\pm$ 0.3} &
    84.8 $\pm$ 0.1 &
    90.0 $\pm$ 0.2 &
    97.1 $\pm$ 0. &
    84.4 $\pm$ 0.2 &
    \textbf{75.5 $\pm$ 0.4} &
    \textbf{96.1 $\pm$ 0.1} &
    \textbf{87.3 $\pm$ 0.2} &
    92.7 $\pm$ 0.3 &
    63.0 $\pm$ 0. \\
    \bottomrule
    \end{tabular}
    }
    \caption{Best transfer accuracies at 80\% and 90\% sparsity for linear and full finetuning, relative to dense transfer. For each downstream task, we present the maximum test accuracy across all sparse methods, highlighting the top accuracy. (We highlight multiple methods when  confidence intervals overlap. Results are averaged across five and three trials for linear and full finetuning, respectively.) Note that in all but three cases (all full finetuning), there is at least one sparse model that is competitive with or better than the dense baseline. }
    \label{tab:accuracy_summary_best}
\end{table*}

\begin{figure}[t]
    \centering
    \includegraphics[width=\linewidth]{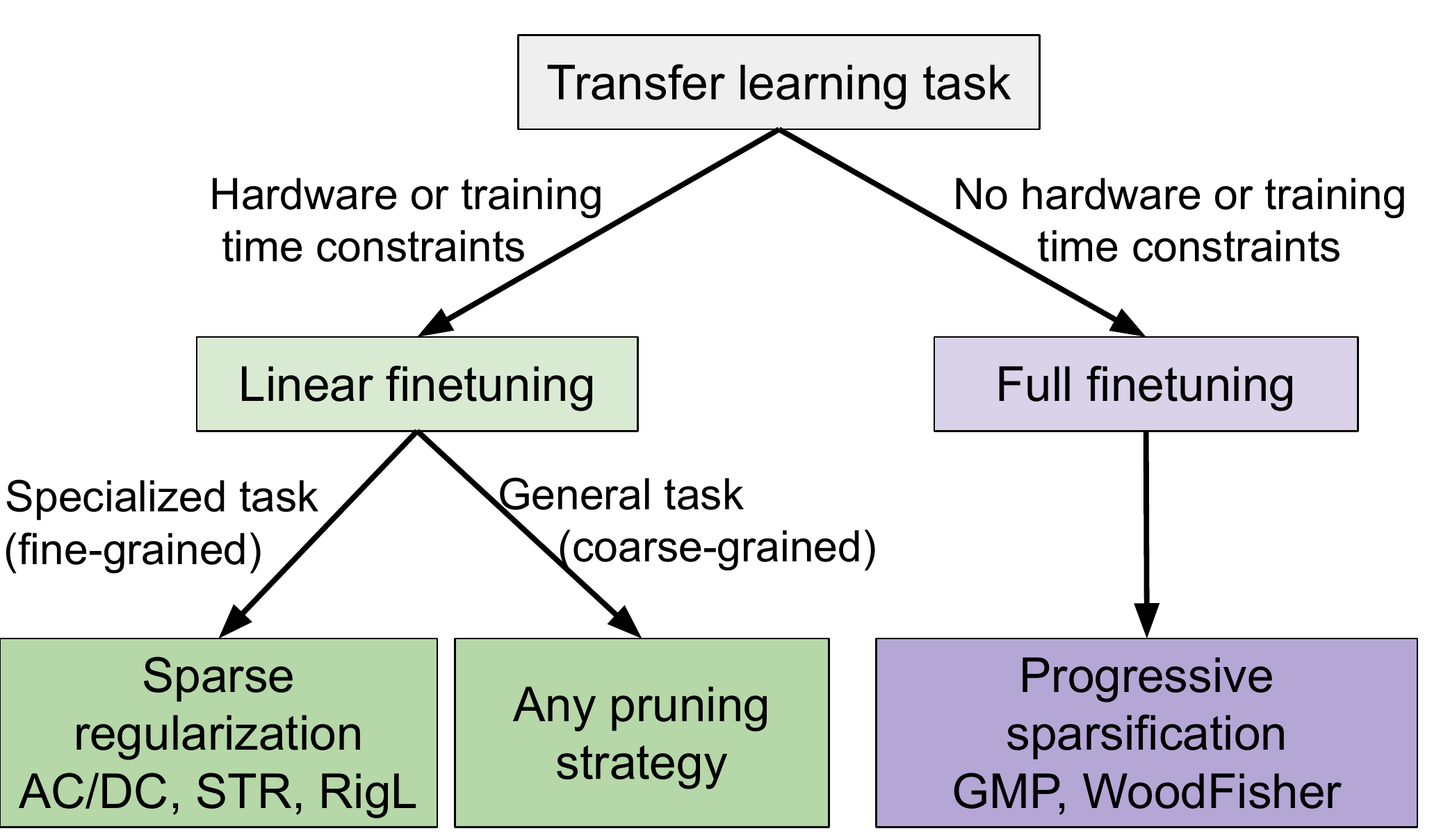}
    \caption{Overview of a suggested decision process when selecting the finetuning and pruning methods to maximize performance and accuracy when doing transfer learning on pruned models.}
    \label{fig:decision_tree}
\end{figure}

We briefly outline our main conclusions, summarized in Figure~\ref{fig:decision_tree} and Table~\ref{tab:accuracy_summary_best}. 
For  \emph{linear finetuning}, sparse models usually match and can slightly outperform dense models. 
     Yet, this is not true for all pruning methods:  \emph{regularization-based} methods perform particularly well, even at high sparsities (e.g. 95\%).  
 For \emph{full finetuning}, which generally provides higher accuracies~\cite{kornblith2019better}, sparse models are also competitive with dense ones, but transfer accuracy is more tightly correlated with accuracy on the ImageNet pre-training task: consequently, less sparse models (e.g. 80\%-90\% sparsity) tend to be more accurate than sparser ones. Moreover, in this setting we find that \emph{progressive sparsification} methods consistently produce models with higher transfer accuracy, relative to regularization methods. We provide a first analysis of this effect, linking it to structural properties of the pruned models.  
 In addition, we observe the markedly lower accuracy of lottery-ticket approaches, especially at the higher levels of sparsity, e.g. $\geq 90\%$, required for computational speedups.   
 
Given the difference in behaviour between linear and full finetuning, we find that there is currently no single ``best'' pruning method for transfer. 
 However, using existing methods, one can consistently achieve order-of-magnitude ($\sim90\%$) compression without loss of accuracy. In turn, these compression levels can lead to  speedups of more than 3$\times$ on sparsity-enabled runtimes. This suggests that sparse transfer may have significant practical potential.

\section{Background and Related Work}

\subsection{Sparsification Techniques}
\label{sec:techniques}

Recently, there has been significant research interest in pruning techniques, and hundreds of different sparsification approaches have been proposed; please see the recent surveys of~\cite{gale2019state} and~\cite{hoefler2021sparsity} for a comprehensive exposition. 
We roughly categorize existing pruning methods as follows. 

 \emph{Progressive Sparsification Methods} start from an accurate \emph{dense} baseline model, and remove weights progressively in several steps, separated by finetuning periods, which are designed to recover accuracy. 
    A classic instance is gradual magnitude pruning (GMP)~\cite{hagiwara1994, han2015learning, zhu2017prune, gale2019state}, which progressively removes weights by their absolute magnitude, measured either globally or per layer. 
    \emph{Second-order} pruning methods, e.g.~\cite{lecun1990optimal, hassibi1993optimal, wang2019eigendamage, singh2020woodfisher, frantar2021efficient} augment this basic metric with second-order information, which can lead to higher accuracy of the resulting pruned models, relative to GMP.  
    
  \emph{Regularization Methods} are usually applied during model training, via sparsity-promoting mechanisms. 
    These mechanisms are very diverse, from surrogates of $\ell_0$ and $\ell_1$-regularization~\cite{yang2019deephoyer, kusupati2020soft}, to variational methods~\cite{molchanov2017variational}, to methods inspired by compressive sensing mechanisms such as Iterative Hard Thresholding (IHT)~\cite{jin2016training, lin2019dynamic, jayakumar2020top, peste2021ac}. We also consider the ``The Rigged Lottery'' (RigL) method ~\cite{evci2020rigging}, which achieves close to state-of-the-art ImageNet results by allowing for dynamic weight pruning and re-introduction with long finetuning periods, to be a regularization method.
    
  \emph{Lottery Ticket Hypothesis (LTH) Methods}~\cite{frankle2018lottery} start from a fully-trained model, and often apply pruning in a single or in multiple incremental steps to obtain a sparse mask over the weights. 
    They then restart training, but restricted to the given mask. 
    Training restarts either from  intialization~\cite{frankle2018lottery}, or by ``rewinding'' to an earlier point during training of the dense model~\cite{frankle2019stabilizing, chen2020lottery, chen2021vision}. (The random mask--initialization combination is thought of as the ``lottery ticket''.)
    Rewinding appears to be required for stable results on large datasets such as ImageNet~\cite{frankle2019stabilizing, chen2020lottery, chen2021lottery}.

The above categorization is clearly approximate: for instance, LTH methods could be viewed as a special case of progressive sparsification, where a specific finetuning approach is applied. Moreover, it is not uncommon to combine approaches, such as regularization and progressive sparsification~\cite{hoefler2021sparsity}. We provide a comparison of the efficacy of these different approaches in the context of transfer learning, by considering multiple methods from each category. To our knowledge, this is the first such detailed study.

Top-1 test accuracy is the standard metric for comparing pruning methods. We also adopt this metric for examining accuracy in the context of transfer, as no such study exists. 
Yet, we wish to highlight recent work \cite{hooker2019compressed, hooker2020characterising, liebenwein_lost_2021} which examines the robustness of pruned models to input perturbations, as well as the impact of pruning on accuracy of specific segments of the data. 

\subsection{Transfer Learning and Sparsity}
 
\paragraph{Dense Transfer Learning.} 
A large body of literature has established that, in general, deep learning architectures transfer well to smaller ``downstream'' tasks, and that full finetuning typically achieves higher accuracy than linear finetuning~\cite{kornblith2019better, salman2020adversarially}. (A very recent study~\cite{kumar2022fine} suggests that this may be inverted on out-of-distribution tasks.) 
These findings extend to related tasks, such as object detection and segmentation~\cite{mensink2021factors}. Kolesnikov et al.~\cite{kolesnikov2020big} have focused on factors determining the success of transfer learning, and on developing reliable fine-tuning recipes. This has been further extended by Djolonga et al.~\cite{djolonga2021robustness}, who concluded that increasing the scale of the original model and dataset significantly improves out-of-distribution and transfer performance, despite having marginal impact on the original accuracy.
Salman et al.~\cite{salman2020adversarially} considered whether \emph{adversarially robust} ImageNet classifiers can outperform standard ones for transfer learning, and find that this can indeed be the case. 
 We complement these studies by examining sparse models and pruning methods. 

\paragraph{Sparse Transfer Learning.} 
One of the earliest works to consider transfer performance for pruned models was~\cite{molchanov2016pruning}, whose goal was to design algorithms which allow the pruning of a (dense) convolutional model when transferring on a target task.  (A similar study  was performed by~\cite{sanh2020movement}  for language models.) 
By contrast, we focus on the different setting where models have already been sparsified on the \emph{upstream} dataset, and observe higher sparsities than the early study of~\cite{molchanov2016pruning}.   

Recent work on sparse transfer learning  has focused specifically on models obtained via the ``Lottery Ticket Hypothesis'' (LTH) approach~\cite{frankle2018lottery}, which roughly states that there exist sparsity masks and initializations which allow accurate sparse networks to be trained \emph{from scratch}. 
There are several works investigating the ``transferrability'' of models obtained via this procedure for different tasks: 
for instance,~\cite{mehta2019sparse} shows that lottery tickets obtained on the CIFAR dataset can transfer well on smaller downstream tasks, while~\cite{chen2020lottery, girish2021lottery} investigate the applicability of lottery tickets for pre-trained language models (BERT), and object recognition tasks, respectively. 
Mallya et al.~\cite{mallya2018piggyback} considered the related but different problem of adapting a fixed network to multiple downstream tasks, by learning task-specific masks.    

The recent work of~\cite{chen2021lottery} considers the transfer performance of LTH for transfer, proposing 
LTH-T, and finding that this method ensures good downstream accuracy at moderate sparsities (e.g., up to 80\%). 
We consider a similar setting, but investigate a wider array of pruning methods (including LTH-T) and additional transfer datasets. 
Specifically, we are the first to compare LTH-T to competitive upstream pruning methods. 
We observe that, on full finetuning, most 
pruning methods consistently outperform LTH-T in terms
of downstream accuracy across sparsity levels, by large margins at high sparsities.

\begin{table}
\centering
\scalebox{0.6}{
\begin{tabular}{@{}ccccc@{}}
\toprule
Dataset & Number of Classes &  Train/Test Examples & Accuracy Metric\\
\midrule
SUN397\cite{xiao2010SUN} & 397 & 19 850 / 19 850 & Top-1 \\
FGVC Aircraft\cite{maji13fgvc-aircraft} & 100 & 6 667 / 3 333 & Mean Per-Class \\
Birdsnap\cite{berg2014birds} & 500 & 32 677 / 8 171 & Top-1 \\
Caltech-101\cite{li2004caltech101} & 101 & 3 030 / 5 647 & Mean Per-Class \\
Caltech-256\cite{holub2006caltech256} & 257 & 15 420 / 15 187 & Mean Per-Class \\
Stanford Cars\cite{krause2013cars} & 196 & 8 144 / 8 041 & Top-1 \\
CIFAR-10\cite{cifar100} & 10 & 50 000 / 10 000 & Top-1 \\
CIFAR-100\cite{cifar100}  & 100 & 50 000 / 10 000 & Top-1 \\
Describable Textures (DTD)\cite{cimpoi2014dtd} & 47 & 3 760 / 1 880 & Top-1 \\
Oxford 102 Flowers\cite{nilsback2006flowers} & 102 & 2 040 / 6 149 & Mean Per-Class \\
Food-101\cite{bossard2014food101} & 101 & 75 750 / 25 250 & Top-1 \\
Oxford-IIIT Pets\cite{parkhi2012apets} & 37 & 3 680 / 3 669 & Mean Per-Class \\
\bottomrule
\end{tabular}
}
\caption{Datasets used as downstream tasks for transfer learning. }
\label{table:datasets}
\end{table}

\section{Sparse Transfer on ImageNet}

\subsection{Experimental Choices}

\paragraph{Transfer Learning Variants.} 
We consider both \emph{full finetuning}, where the entire set of features is optimized over the downstream dataset, and \emph{linear finetuning}, where only the last layer classifier is finetuned, over sparse models. 
In the former case, with the exception of the final classification layer and the batch normalization (BN) parameters, only the nonzero weights of the original model are optimized, and the mask is kept fixed. 

We do not consider from-scratch training and pruning on the downstream task, for two reasons. 
First, from-scratch training is often less accurate than (dense) transfer learning in the same setting~\cite{kornblith2019better, mensink2021factors}. 
As our experiments will show, transfer from \emph{sparse models} can often match or even slightly outperform transfer from \emph{dense models}. 
Second, since training from scratch is typically less accurate than transfer~\cite{kornblith2019better}, it seems unlikely that training and pruning from scratch will outperform sparse transfer. We give  evidence for this claim in Appendix~\ref{appendix:transfer-better-than-scratch}.
One practical advantage of this approach is not needing hyper-parameter tuning with respect to compression on the downstream dataset. 

\paragraph{Network Architectures.} 
Our study is based on an in-depth analysis of sparse transfer using the ResNet50  architecture~\cite{he2016deep}. 
This architecture has widespread practical adoption, and has been extensively studied in the context of transfer learning~\cite{kornblith2019better, salman2020adversarially}.  
Importantly, its compressibility  has also emerged as a consistent benchmark for CNN pruning methods~\cite{hoefler2021sparsity}. 
We further validate some of our findings on ResNet18, ResNet34 and MobileNet~\cite{howard2017mobilenets} architectures. 
In addition, we investigate transfer between two classical object detection tasks, MS COCO~\cite{lin2014microsoft} and Pascal VOC~\cite{everingham2010pascal}, using variants of the  YOLOv3 architecture~\cite{redmon2018yolov3}.

\paragraph{Sparsification Methods.} 
For our study, we chose the pruning methods providing top validation accuracy for each method type in Section~\ref{sec:techniques}. 
For \emph{progressive sparsification methods}, we use the leading WoodFisher~\cite{singh2020woodfisher} and Gradual Magnitude Pruning (GMP)~\cite{hagiwara1994, han2015learning, zhu2017prune, gale2019state} methods. 
For \emph{regularization methods}, we consider the leading Soft Threshold Weight Reparametrization (STR) \cite{kusupati2020soft}, and Alternating Compression/Decompression (AC/DC)\cite{peste2021ac} methods. Additionally, we include the ``The Rigged Lottery'' (RigL) method \cite{evci2020rigging} with Erd{\H o}s-R{\' e}nyi-Kernel (ERK) weight density. Compared to STR and AC/DC, RigL extends the training schedules on ImageNet by up to 5x, and does \emph{sparse training} for most of the optimization steps. We consider both the standard version or RigL (RigL ERK 1x), and the variant with 5x training iterations (RigL ERK 5x). Finally, for \emph{LTH Methods}, we consider the LTH-for-Transfer (LTH-T) method of~\cite{chen2021lottery}, which precisely matches our setting.  In this version, the authors apply the masks obtained through progressive sparsification methods 
directly to the original trained ImageNet dense model, and evaluate the transfer accuracy of this masked model through full finetuning on different downstream tasks. 

We focus on \emph{unstructured} pruning, as these methods are the most studied in the pruning literature, have well established benchmarks, and achieve the best trade-off between accuracy and compression. We include results for full finetuning from models with structured sparsity in Appendix \ref{appendix:structured_sparsity}, showing that, given a fixed accuracy level upstream, structured-sparse models tend to underperform unstructured-sparse models for transfer.

When available, we use original sparse PyTorch checkpoints, and the exact architectures used by the upstream models. However, since the STR and RigL models were trained using label smoothing, which has been shown in \cite{kornblith2019better} to decrease transfer accuracy, we used retrained versions of these models on ImageNet, without label smoothing. The results we discuss in the following sections are for these versions, which indeed perform better, particularly on linear finetuning (see Appendix \ref{section:label_smoothing}). We manually ported RigL checkpoints from TensorFlow to PyTorch  (see Table \ref{table:sparse_eval} for all ImageNet results).

\begin{table}
\centering
\scalebox{0.7}{%
\begin{tabular}{@{}cccccccc@{}}
\toprule
 Sparsity  & 
Method & 
\multicolumn{1}{c}{\begin{tabular}[c]{@{}c@{}} Original \\ Validation \end{tabular}} & \multicolumn{1}{c}{\begin{tabular}[c]{@{}c@{}} Reassessed \\ Labels \end{tabular}} & \multicolumn{1}{c}{\begin{tabular}[c]{@{}c@{}} ImageNetV2 \\ (Average) \end{tabular}} & \\
\toprule
$0\%$ & Dense & 76.8\% & 83.1\%  & 72\%\\

\midrule
$80\%$ & AC/DC & 76.2\% & 82.9\% & 71.8\%  \\
& STR & 75.5\% & 81.9\% & 70.3\% \\
& WoodFisher & \textbf{76.7\%} & \textbf{83.2\%} & \textbf{72.3\%} \\
& GMP & 76.4\% & 82.9\% & 71.6\% \\
& RigL ERK 1x & 74.8\% & 81.3\% & 70.2\% \\
& RigL ERK 5x & 75.8\% & 81.6\% & 70.6\% \\

\midrule
$90\%$ & AC/DC & 75.2\% & 82.2\% & 70.6\% \\
& STR & 74.0\% & 80.9\% & 69.1\% \\ 
& WoodFisher & 75.1\% & \textbf{82.4\%} & \textbf{71.1\%}\\
& GMP & 74.7\% & 81.6\% & 70.1\% \\
& RigL ERK 1x & 73.2\% & 80.0\% & 67.9\% \\
& RigL ERK 5x & \textbf{75.7\%} & 81.9\% & 70.6\%\\

\midrule
$95\%$ & AC/DC & 73.1\% & 80.4\% &  68.6\% \\ 
& STR & 70.4\% & 77.9\% & 66.0\% \\
& WoodFisher & 72.0\% & 79.8\% & 67.6\% \\ 
& RigL ERK 1x & 70.1\% & 77.5\% & 65.5\% \\
& RigL ERK 5x &\textbf{ 74.0\%} & \textbf{80.8\%} & \textbf{69.0\%} \\

\bottomrule
\end{tabular}
}
\caption{Accuracy of the pruning methods we use, at different sparsity levels, evaluated on different ImageNet validation sets.} 
\label{table:sparse_eval}
\end{table}

\paragraph{Downstream tasks and training.}
We follow~\cite{salman2020adversarially} in using the twelve standard transfer benchmark datasets described in Table \ref{table:datasets}, 
which span several domains and sizes. 
We transfer all parameters of the upstream model except for the last (fully connected) layer, which is adjusted to the number of classes in the downstream task, using Kaiming uniform initialization \cite{he2015delving}, and kept dense. 
This may slightly change the sparsity of the model, as in some cases the final layer was sparse. As a convention, when discussing sparsity levels, we refer to the \emph{upstream} checkpoint sparsity. We provide   full training hyperparameters in  Appendix~\ref{appendix:hyperparams}.

\paragraph{Performance metrics.} The main quantity of interest is the top-1 validation accuracy on each transfer task, measured for all the pruned models, as well as for the dense baselines. In some cases, we use the mean per-class validation accuracy following the convention for that dataset (see Table \ref{table:datasets}).
To determine the overall ``transfer potential'' for each pruning method, we further present the results aggregated over the downstream tasks. Since datasets we use for transfer learning have varying levels of difficulty, as reflected by the wide range of transfer accuracies, we compute for each downstream task and model the \emph{relative increase in error} over the dense baseline. Specifically, if $B$ is the baseline dense model, then for every downstream task $D$ and sparse model $S$ we define the relative increase in error as $\alpha_{D, S} = \frac{err_{D, S} - err_{D, B}}{err_{D, B}}$, where $err_{D, S}$ is the error corresponding to the top validation accuracy for model $S$ trained on dataset $D$. For each pruning method and sparsity level, we report the mean and standard error of $\alpha_{D, S}$, computed over all downstream tasks. 

We also examine the computational speedup potential of each method, along with its accuracy. 
For \emph{inference-time} speedups, our findings are in line with 
previous work, e.g.~\cite{elsen2020fast, singh2020woodfisher, peste2021ac}. We will therefore focus on the \emph{training-time} speedup potential in the case of \emph{linear finetuning}, which are usually close to inference-time speedups, as the only difference is the training time of the classifier layer.   

\subsection{Validation Accuracy on ImageNet Variants}
\label{sec:validation_results}

To set a baseline, we first examine accuracies on the original ImageNet validation set, and on  different versions of this validation set.  
Namely, we use the ImageNet ``reassessed labels'' \cite{beyer2020we}, where the original ImageNet validation images are re-assessed by human annotators. We also use three different ImageNetV2 validation sets \cite{recht2019imagenet}, where the new images with a similar data distribution are gathered based on different criteria. We report the average ImageNetV2 accuracy across these three variants in Table \ref{table:sparse_eval}. 

\paragraph{Discussion.} We observe that RigL ERK 5x  outperforms all methods on the original validation set at 90\% and 95\% sparsity, followed by AC/DC, GMP and WoodFisher. At 80\% sparsity, WoodFisher has the best original validation accuracy, followed closely by GMP and AC/DC. However, despite the gap in original validation accuracy between RigL ERK 5x and other methods, the results on new variants of the validation set still reveal some interesting patterns. For example, WoodFisher outperforms all methods at 80\% and 90\% sparsity on the reassessed labels, followed closely by AC/DC. This is true also for ImageNetV2, where WoodFisher outperforms all methods at 80\% and 90\% sparsity. At 95\% sparsity, however, RigL ERK 5x outperforms all methods considered, including on the reassessed labels and ImageNetV2, and is followed by AC/DC. 
Generally, the accuracies on the reassessed labels and ImageNetV2 correlate well with those on the original images, which suggests that top performing methods can ``extrapolate'' well. 

\subsection{Linear Finetuning}
\label{subsec:linear-finetuning}

Next, we study the transfer performance of different types of pruning methods in the scenario where only the linear classifier ``on top'' of a fixed representation is trained on the downstream task. Specifically, we study the simple setup where the features prior to the final classification layer of the pre-trained model are extracted for all samples in the transfer dataset and stored into memory for use when training the downstream linear classifier. 
Although this approach typically results in lower accuracy relative to full finetuning~\cite{kornblith2019better, salman2020adversarially}, it has significant practical advantages. Specifically, the features can be precomputed, which eliminates the forward passes through the pretrained network. In this setup, we do not apply any data augmentation on the transfer samples and we use the Batch Normalization statistics of the pretrained network on ImageNet.

We optimize the linear classifier using SGD with momentum, weight decay and learning rate annealing, following~\cite{salman2020adversarially}. (The results are typically well-correlated with those obtained when using data augmentation during training, or using different optimizers~\cite{kornblith2019better}). 
In Section~\ref{subsec:speedup}, we show that training speed-ups can also be obtained in an online learning setup, where new samples are executed through the backbone network, by taking advantage of the backbone sparsity.

\begin{figure*}[t]
    \centering
    \includegraphics[width=\textwidth]{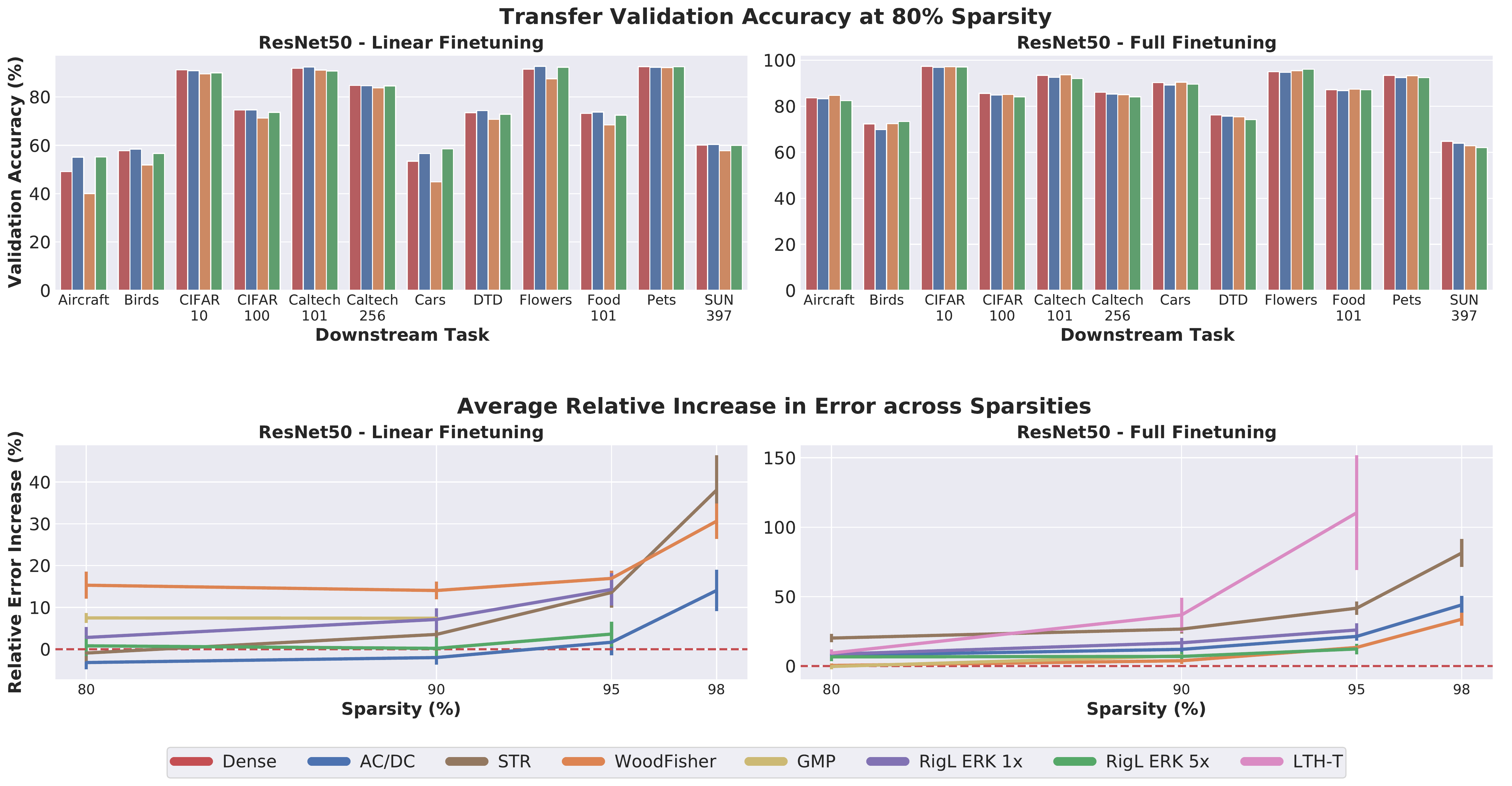}
    \caption{(top row) Validation accuracy for selected pruning strategies at 80\% sparsity. (bottom row) Average increase in validation error relative to the dense baseline; lower values are better. Best viewed in color. }
    \label{fig:megaplot}
\end{figure*}

The results for linear finetuning are shown in Figure \ref{fig:megaplot}, and Appendix Table \ref{table:rn50_linear_all}. We exclude the LTH-T method from this analysis, as it is designed for full finetuning, and its transfer accuracy in the linear scenario is indeed very low (see Appendix Table \ref{table:rn50_linear_all}).

Overall, the results clearly show that the choice of pruning strategy on the upstream task can result in significant differences in performance on downstream tasks. These differences are more apparent for specialized downstream tasks, with fine-grained classes. For example, consider Aircraft, where for 80\% sparse models we see a 15\% gap in top-1 test accuracy between the best-performing sparse models (AC/DC and RigL, 55\%) and the worst-performing one (WoodFisher, 40\%). 

\begin{figure}
    \centering
    \includegraphics[width=\linewidth]{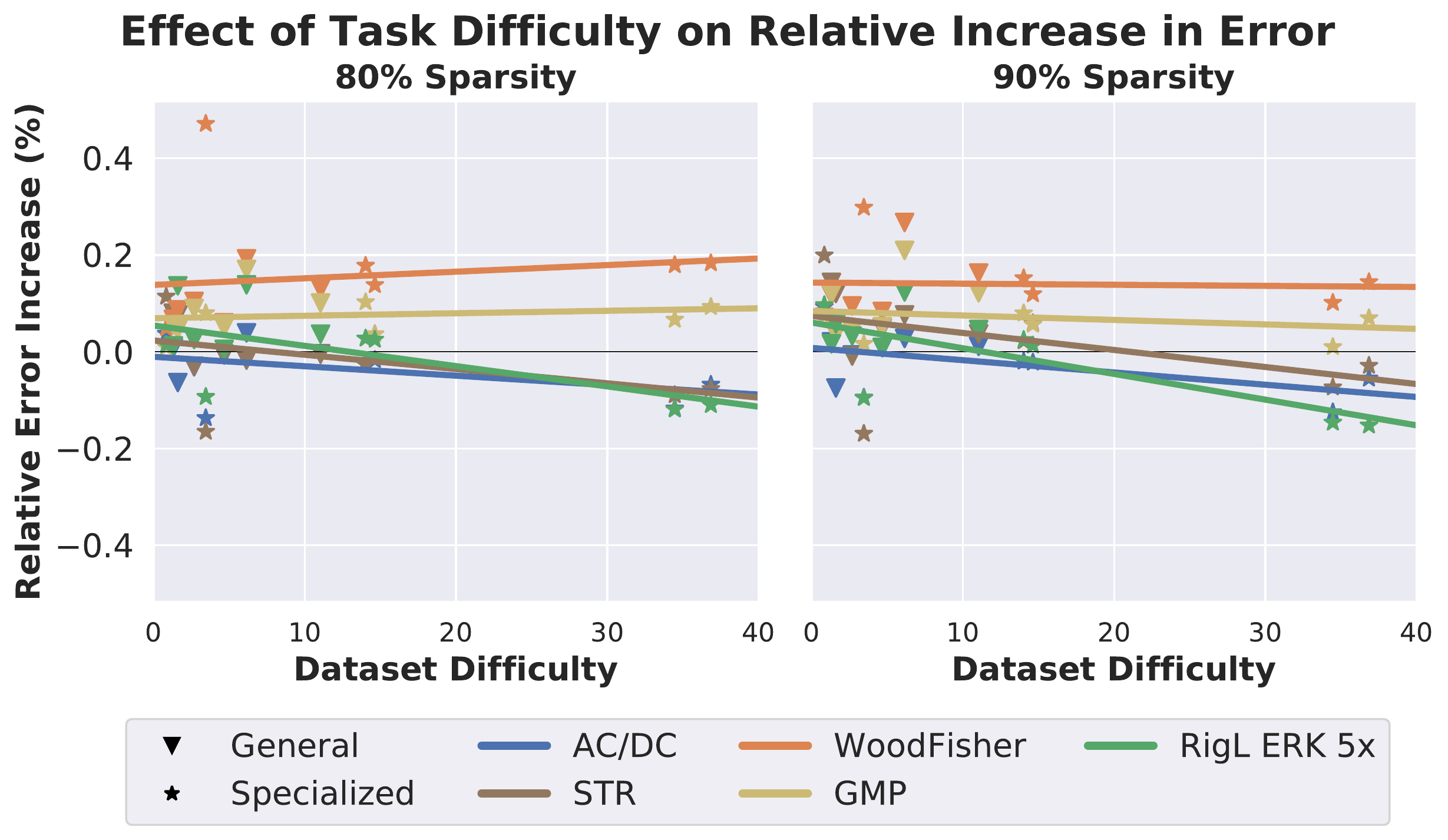}
    \caption{Effect of task difficulty on various pruning strategies for transfer with linear finetuning. Best viewed in color.} 
    \label{fig:difficulty_improvement}
\end{figure}

Following this observation, we study the correlation between the downstream task difficulty and relative increase in error for different pruning strategies. For this purpose, we use the difference in top-1 validation accuracy between full and linear finetuning on the dense backbone as a proxy for the difficulty of a downstream task. Intuitively, a small gap between full and linear finetuning would suggest that the upstream features are directly transferable, and thus the downstream task can be considered ``easy''. Conversely, a large gap would indicate that the pre-trained features are not enough to capture the internal representation of the data, making the downstream task more ``difficult''. Additionally,  we categorize the downstream tasks into \emph{general} (Caltech-101/256, CIFAR-10/100, DTD, SUN397) vs. \emph{specialized} (Aircraft, Birds, Cars, Flowers, Food-101, Pets); this is similar to previous work \cite{kornblith2019better}. Figure~\ref{fig:difficulty_improvement} suggests that specialized datasets tend to have higher difficulty scores. 

Following this definition and categorization, we measure, for each pruning strategy, the relative error increase over the dense model against the task difficulty. Figure~\ref{fig:difficulty_improvement} shows the behavior for all pruning methods considered at 80\% and 90\% sparsity. Interestingly, we observe a trend for \emph{regularization methods} (AC/DC, STR, RigL) to improve over the dense baseline with increased task difficulty, which is more apparent at higher sparsity (90\%). In contrast, progressive sparsification methods (GMP, WoodFisher) do not show a similar behavior. This suggests that regularization pruning methods are a better choice for linear transfer (sometimes even surpassing the dense performance) when the downstream task is more specialized or more difficult.

Another particularity of linear finetuning from sparse models is that the sparsity level is not highly correlated with the performance on the downstream tasks. This is apparent, for example, for AC/DC and RigL, where, despite the 1-2\% gap in ImageNet accuracy between the 80\% and 90\% sparse models, the relative error with respect to the dense baseline stays quite flat. A similar trend can be observed for other pruning methods as well. However, extremely sparse models (98\%) tend to perform worse, probably due to feature removal and degradation. 

In summary, we observe that 1) some sparsification methods can consistently match or even sometimes outperform dense models; 2) there is a correlation between transfer performance for regularization-based methods and downstream task difficulty; and 3) higher sparsity is not necessarily a disadvantage for transfer performance. 

\subsection{Full Finetuning}
\label{subsec:full-finetuning}

We now consider the  \emph{full finetuning} scenario. Here, we re-initialize the final classification layer and fix it as dense, then finetune the unpruned weights so that  the network is sparse throughout training. The results are summarized in  Figure \ref{fig:megaplot}, and detailed in Appendix Table \ref{table:rn50_full_all}. 

Similar to linear finetuning, we see substantial performance variations among pruning strategies when transferred to the downstream tasks. Typically, progressive sparsification methods (WoodFisher, GMP) tend to transfer better than regularization and lottery ticket methods. The differences in test accuracy, measured at the same sparsity level, are typically small, on the order of 1--3\%;
the exception is LTH-T, which is competitive at low sparsity (80\%) but incurs severe accuracy drops at sparsities $\geq 90\%$.  

In contrast to linear finetuning, we see a consistent trend of decreasing quality with increased sparsity. This is not surprising, since full finetuning can take advantage of the additional parameters available in denser models to better fit the downstream data. Nevertheless, \emph{progressive sparsification} methods (GMP and WoodFisher) result in downstream performance nearly on par with dense models at 80\% and 90\% sparsity. These methods show better performance than regularization-based methods (AC/DC, STR, and RigL), a direct reversal of the results of linear finetuning. 

For specific downstream tasks, however, there is considerable variability---while WoodFisher and GMP are consistently the top or near-top performing models across all tasks, other methods show considerable task dependence. For instance, while AC/DC is the top performing method across different sparsities for three of the twelve tasks (SUN397, Caltech-256, and DTD), it shows a considerable gap compared to the best-performing methods on Aircraft, Cars, and CIFAR-10. 
Generally, STR performs  worse  on  full  finetuning,  compared  to other regularization methods. Furthermore, RigL ERK 1x performs roughly on par with AC/DC, despite having a lower validation accuracy on ImageNet; however, the extended training of RigL ERK 5x gives the tansfer accuracies a considerable boost, putting RigL ERK 5x almost on par with WoodFisher, especially at higher sparsities. This finding opens the intriguing possibility that \emph{extended} training may be beneficial for pruning methods in the full finetuning regime. We present further evidence in favor of this hypothesis in Appendix~\ref{appendix:extended-training}; namely, we additionally show that extending the training time of upstream AC/DC models (3x or 5x) significantly improves their transfer performance on all downstream tasks. 
Finally, LTH-T shows fairly competitive performance at 80\% sparsity, but its transfer accuracy declines dramatically on six of the twelve datasets (SUN397, Caltech-101, Caltech-256, DTD, Flowers, and Pets) as sparsity increases. 
Since the LTH-T model relies mainly on transferring the sparsity mask across tasks, this suggests that the additional information present in the \emph{weights}, leveraged by other methods, may be beneficial. 

In sum, if the goal is to perform full finetuning on downstream tasks, then \emph{progressive sparsification} methods are a good choice. 
They consistently outperform regularization methods across a wide range of tasks, and offer comparable performance to the dense backbone at 80\% and 90\% sparsity.

\subsection{Discussion} 
\label{subsec:discussion_filters}

The results of the last two sections show an intriguing performance gap between pruning methods, depending on the transfer approach. Investigating further, we examine the sparse structure of the resulting pruned models, by measuring the percentage of convolutional filters that are completely pruned away during the training phase of the sparse ResNet50 backbones on the original ImageNet dataset. 
The results in Table~\ref{table:empty-filters} show the differences in the number of zeroed-out channels among pruning methods. 
We observe that AC/DC has a large number of channels that are fully removed during ImageNet training and pruning, on average 2-4 more at 80\% and 90\% sparsity, compared to other models; this results in fewer features that can be trained during full finetuning. By contrast, the sparsity in GMP and WoodFisher is less structured and thus can express additional features, which can be leveraged during finetuning. We further illustrate this effect in Appendix \ref{appendix:structured_sparsity}, where we fully finetune from models with structured sparsity.

\begin{table}[h]
\centering
\scalebox{0.65}{%
\begin{tabular}{ccccccc}
\toprule
\multicolumn{1}{c}{\begin{tabular}[c]{@{}c@{}} Pruned  \\ Filters (\%) \end{tabular}} & {AC/DC} & {WoodFisher} & {GMP} & {STR} & \multicolumn{1}{c}{\begin{tabular}[c]{@{}c@{}} RigL ERK\\ 1x \end{tabular}}& \multicolumn{1}{c}{\begin{tabular}[c]{@{}c@{}} RigL ERK\\ 5x \end{tabular}} \\
\midrule
80\% & \bf{2.9\%} & 0.9\% & 1.6\% & 0.5\% & 0.2\% & 0.6\% \\ 
90\% & \bf{8.5\%} & 2.0\% & 2.8\% & 2.0\% & 1.2\% & 2.7\% \\
95\% & \bf{18\%} & 3.0\% & - & 6.0\% & 4.3\% & 9.1\% \\ 
\bottomrule
\end{tabular}
}
\caption{Percentage of convolutional filters that are completely masked out, for different pruning methods on ResNet50, at different sparsity levels. AC/DC has significantly more pruned filters.}
\label{table:empty-filters}
\end{table}

In the case of linear finetuning, we hypothesize that the accuracy reversal in favor of AC/DC can be attributed to a regularizing effect, which produces more ``robust'' features. The same effect appears to be present in RigL ERK 5x at 95\% sparsity, which also has significantly many fully-pruned  filters.

\subsection{Training Speed-Up using Linear Finetuning}
\label{subsec:speedup}

One of the main benefits of sparse models is that they can provide \emph{inference speed-ups} when executed on sparsity-aware runtimes~\cite{elsen2020fast, singh2020woodfisher, peste2021ac, pmlr-v119-kurtz20a}. 
For linear finetuning, this can also imply \emph{training time} speed-ups, since the sparse backbone is fixed, and only used for inference. 
We illustrate this in an ``online learning'' setting, where training samples arrive dynamically at the device. We first compute the corresponding features using the sparse backbone. 
Then, we use these features to train the linear classifier. Thus, the forward pass can benefit from speed-ups due to sparsity. 

\begin{figure}[t]
    \centering
    \includegraphics[height=1.8in]{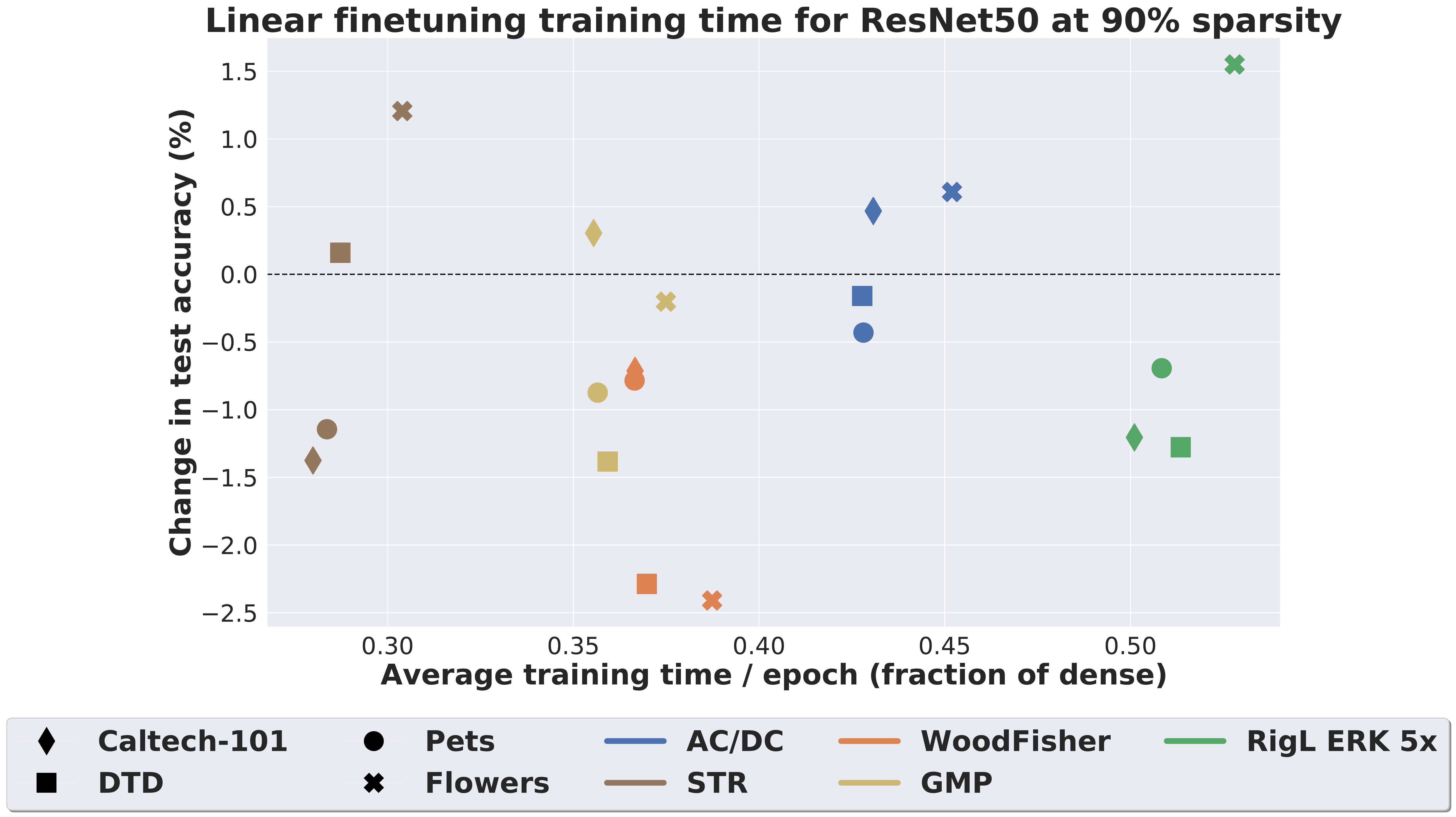}
    \caption{Average epoch time vs. gap in validation accuracy, compared to the dense baseline. Results are shown for four different downstream tasks, using linear finetuning from ResNet50 90\% sparse models. Lower is better; best viewed in color.}
    \label{fig:time-vs-acc}
\end{figure}

To measure these effects, we integrated the freely-available sparsity-aware DeepSparse CPU inference engine~\cite{pmlr-v119-kurtz20a, NM} into our PyTorch pipeline. Specifically, we use sparse inference for online feature extraction. 
We report overall training speedup, in terms of average training time per epoch on the downstream task, divided by the average training time using the dense baseline. 
We use batch size 64 and data augmentation; otherwise, hyperparameters are identical to the linear finetuning experiment in Section~\ref{subsec:linear-finetuning}. 
We execute on an Intel E5-1650 CPU with 12 cores, which would be similar in performance to a recent laptop CPU.
The speed-ups we report are proportional to the inference speed-ups of the respective sparse backbone models. The only difference is the cost of optimizing the last layer, which varies in size with the number of classes. 

Figure~\ref{fig:time-vs-acc} shows results on four downstream tasks, Pets, Flowers, DTD and Caltech-101, where the backbone ResNet50 models have 90\% sparsity. 
We report the training speed-up vs. the difference in validation accuracy, compared to the dense baseline. Furthermore, we show speed-up numbers for additional sparsity levels (80\% and 95\%), in Table~\ref{table:train-speed}; these numbers representing the average training time per epoch are computed on the Caltech-101 dataset, but the speed-up factor should be similar for other datasets as well, since the training time per batch is almost proportional to the inference through the backbone network. 
Results show that \emph{using sparse backbones can reduce training time by 2-4x for linear transfer, without negative impact on validation accuracy}. 

\begin{table}[h]
\centering
\scalebox{0.9}{%
\begin{tabular}{cccccc}
\toprule
{Sparsity} & {STR} & {GMP} & {WoodFisher} & {AC/DC} & {RigL 5x} \\
\midrule
80\% & $0.44\times$ & $0.50\times$ & $0.53\times$ & $0.60\times$ & $0.71\times$ \\
90\% & $0.28\times$ & $0.36\times$ & $0.37\times$ & $0.43\times$ & $0.50\times$ \\
95\% & $0.22\times$ & {N \slash A} & $0.28\times$ & $0.32\times$ & $0.36\times$ \\
\bottomrule
\end{tabular}
}
\caption{Average training time per epoch for linear finetuning using sparse models, as a fraction of the time per epoch required for the dense backbone. The numbers shown are computed on the Caltech-101 dataset.}
\label{table:train-speed}
\end{table}

\section{Extensions}
\label{subsec:additional_arch}

\paragraph{ResNet18/34 and MobileNet Experiments.} 
We have also executed a subset of the experiments for ResNet18, ResNet34 and MobileNetV1~\cite{howard2017mobilenets} models trained on ImageNet. 
These results largely validate our analysis above, and are therefore deferred to Appendix~\ref{appendix:rn18_rn34} and~\ref{appendix:mobilenet}. 
Specifically, regularization-based methods also match or slightly outperform dense ones on linear transfer.
However, for MobileNetV1, we observe that sparse models can match the dense baseline transfer performance only at lower sparsities (up to 75\%), probably due to the lower parameter count.  

\paragraph{Structured Sparsity Experiments.} We also performed full finetuning using models with \emph{structured sparsity}, for both ResNet50 and MobileNet. Our findings, presented in Appendix~\ref{appendix:structured_sparsity}, show that structured-sparse models tend to transfer worse compared to unstructured methods. 

\paragraph{Sparse Transfer using YOLO.} 
We also examined transfer performance between YOLO V3~\cite{redmon2018yolov3} and YOLO ``V5''~\cite{ultralytics} models for object detection, 
trained and pruned on the COCO dataset~\cite{lin2014microsoft}, which are then transferred to the VOC dataset~\cite{everingham2010pascal} using full finetuning. 
Table~\ref{table:all_yolo} presents results in terms of mean Average Precision (mAP@0.5). 
Results show a strong correlation between accuracy on the original COCO dataset and that on VOC, confirming our claims. 
We observed similar trends in a segmentation setup, which we cover in Appendix~\ref{appendix:segmentation-sparse}.

\begin{table}[t!]

\centering
\scalebox{0.7}{%
\begin{tabular}{cccc}
\toprule
{Architecture} & {YOLOv3} & {YOLOv5S} & {YOLOv5L} \\
{Pruning} & {90\% Sparsity} & {75\% Sparsity} & {85\% Sparsity} \\
\midrule
COCO Dense & 64.2  & 55.6 & 65.4 \\
COCO Pruned & 62.4  & 53.4 & 64.3 \\
\midrule
VOC Dense Transfer & 86.0  & 83.73 & 90.0 \\
VOC Pruned Transfer & 84.0  & 81.72 & 89.35 \\
\bottomrule
\end{tabular}
}
\caption{Accuracies for Sparse Transfer from COCO to VOC.}
\label{table:all_yolo}
\end{table}

\section{Conclusions and Future Work}

We performed an in-depth study of the transfer performance of sparse models, and showed that pruning methods with similar accuracy on ImageNet can have surprisingly disparate Top-1 accuracy when used for transfer learning. In particular, regularization-based methods perform best for linear finetuning; conversely, progressive sparsification methods such as GMP and WoodFisher tend to work best when full finetuning is used. 
One limitation of our study is that it only investigated accuracy as a measure of performance for transfer learning tasks. Additional research is needed towards designing pruning strategies with good performance across \emph{both} linear and full finetuning, and towards considering metrics past Top-1 accuracy, such as bias and robustness. 
Another limitation is that we considered a (standard) fixed set of transfer datasets; our study should be extended to other, more complex transfer learning scenarios, such as distributional shift~\cite{koh2021wilds}. 
Further investigation could also systematically examine other types of compression, such as quantization and structured pruning, potentially in conjunction with unstructured pruning, which was the focus of our current study. 
Other interesting areas for future work would be understanding the performance gap between full finetuning and linear finetuning, and realizing training speedups for sparse full finetuning, by taking advantage of the fixed sparsity in the trained model. 

\section*{Acknowledgments}

The authors would like to sincerely thank Christoph Lampert and Nir Shavit for  fruitful discussions during the development of this work, Eldar Kurtic for experimental support, Utku Evci for providing RigL models, and the authors of \cite{chen2021lottery} for sharing the LTH-T masks. 
EI was supported in part by the FWF DK VGSCO, grant agreement number W1260-N35, while AP and DA acknowledge generous support by the ERC, via Starting Grant 805223 ScaleML.

{\small
\bibliographystyle{ieee_fullname}
\bibliography{references.bib}
}

\appendix
\onecolumn
\section*{Appendix}

\setcounter{table}{0}
\renewcommand{\thetable}{\Alph{section}.\arabic{table}}
\setcounter{figure}{0}
\renewcommand{\thefigure}{\Alph{section}.\arabic{figure}}

\section{Comparison between From-Scratch Pruning and Transfer}
\label{appendix:transfer-better-than-scratch}

We now present an experiment which supports our claim that from-scratch pruning and finetuning is inferior to transfer from ImageNet sparse models. For instance, image classification on the CIFAR-100 \cite{cifar100} dataset using a WideResNet \cite{zagoruyko2016wide} architecture, following~\cite{peste2021ac}, the AC/DC and GMP pruning methods at 90\% sparsity reach 79.1\% and 77.7\% Top-1 validation accuracies, respectively. In contrast, finetuning from a ResNet50 backbone pruned on ImageNet using AC/DC and GMP at 90\% sparsity reaches a validation accuracy of 83.9\% and 84.4\%, respectively (please see Table~\ref{table:rn50_full_all} for the results). This example serves to illustrate the  significant accuracy gains from using transfer learning with sparse models, as opposed to training sparse models from scratch. 

\section{Hyperparameters and Training Setup}
\label{appendix:hyperparams}

Here we discuss the general hyperparameters and experimental setup used for the full and linear finetuning experiments. 
Regarding data loading image augmentation settings, we are careful to match them to the ones used in the original upstream training protocol. Specifically, this affects the choice of whether to use Bicubic or Bilinear image interpolation for image resizing; for example, RigL models were trained using Bicubic interpolation, whereas the other pruning methods considered used the Biliniar interpolation. All ResNet and MobileNet models considered were trained using standard ImageNet-specific values for the normalization mean and standard deviation. In the case of full finetuning, we used dataset-specific normalization values for the downstream tasks; these were obtained by loading the dataset once with standard data augmentations and computing the means and variances of the resulting data. For linear finetuning, we use center cropping of the images, followed by normalization using standard ImageNet values. 
For both full and linear finetuning, we use the same training hyperparameters as \cite{salman2020adversarially}; specifically, we train for 150 epochs, decreasing the initial learning rate by a factor of 10 every 50 epochs. We use 0.01 as the initial learning rate for all linear finetuning experiments; for full finetuning, we empirically found 0.001 to be the initial learning rate which gives comparable results for most datasets except Aircraft and Cars, for which we use 0.01.
Our experiments were conducted using PyTorch 1.8.1 and NVIDIA GPUs. 
All full finetuning experiments on the ResNet50 backbone were repeated three times and all linear finetuning experiments five times.


\section{Extended ResNet50 Results}
\label{appendix:rn50-results}

In this section, we provide additional details, together with the complete results for our experiments for linear and full finetuning from ResNet50, presented in Sections~\ref{subsec:linear-finetuning} and ~\ref{subsec:full-finetuning}. For each pruning method, we used a range of sparsity levels, and trained linear and full finetuning for each model and sparsity level, on all 12 downstream tasks; each experiment was repeated 5 times for linear and 3 times for full finetuning. Note that checkpoints for some pruning methods were not available for some of the higher sparsities.

\subsection{Linear Finetuning Results}
\label{appendix:rn50-linear}

\begin{figure}[h]
    \centering
    \includegraphics[width=0.9\textwidth]{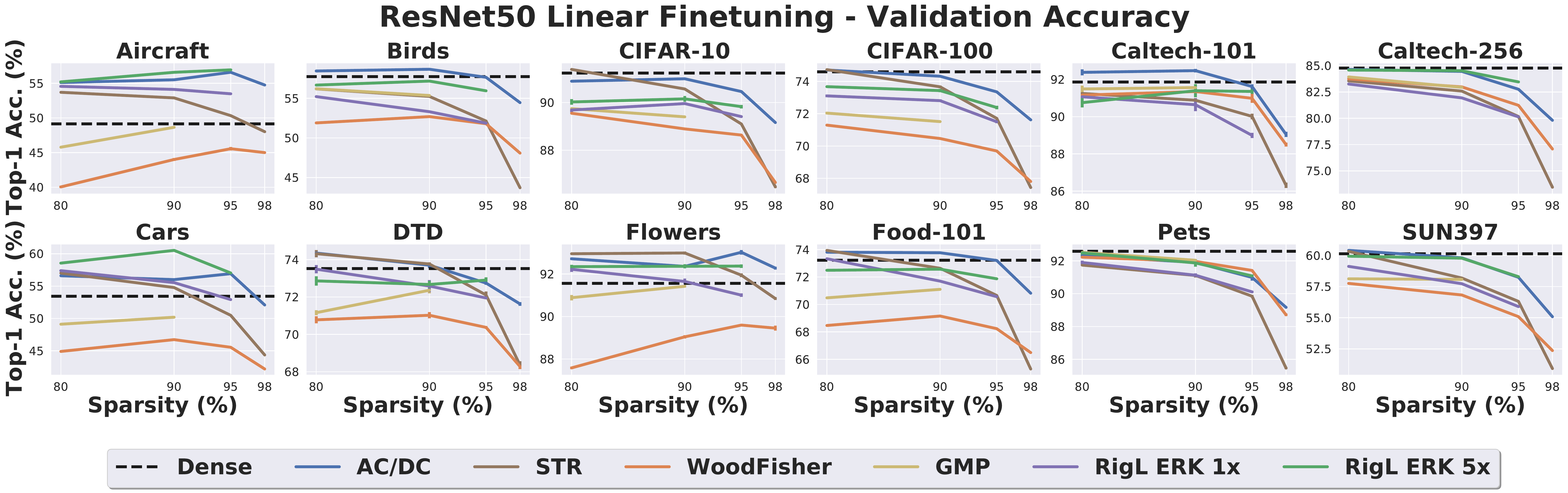}
    \caption{(ResNet50) Per-dataset downstream validation accuracy for transfer learining with \emph{linear finetuning}.}
    \label{fig:individual_resnet_linear}
\end{figure}

We provide the complete results for our linear finetuning experiments on each downstream task, for all pruning methods and sparsity levels considered. The results for the transfer accuracies for each pruning strategy, sparsity level, and downstream task are presented in Figure~\ref{fig:individual_resnet_linear} and Table~\ref{table:rn50_linear_all}. We discussed in Section \ref{subsec:linear-finetuning} that regularization methods match and even sometimes outperform the dense baseline transfer performance. Note that this fact is valid not only in aggregate, but also at the level of each individual dataset. 

In table \ref{table:rn50_linear_all}, we also include the linear transfer results for LTH-T. We note that the generally poor performance of the method, especially for more specialized tasks and higher sparsity levels, should not be taken as a criticism of the method itself: this use case is clearly contrary to the method's design, and the spirit of the original Lottery Ticket Hypothesis (which aims to discover masks with the intent to retrain, rather than final weights). Rather, we include these results to provide quantitative justification for the omission of LTH-T from any further analyses, and supporting the original authors' point that additional finetuning is \emph{necessary} in order to obtain a competitive lottery ticket for transfer learning.

\begin{figure}[h]
    \centering
    \includegraphics[width=0.9\textwidth]{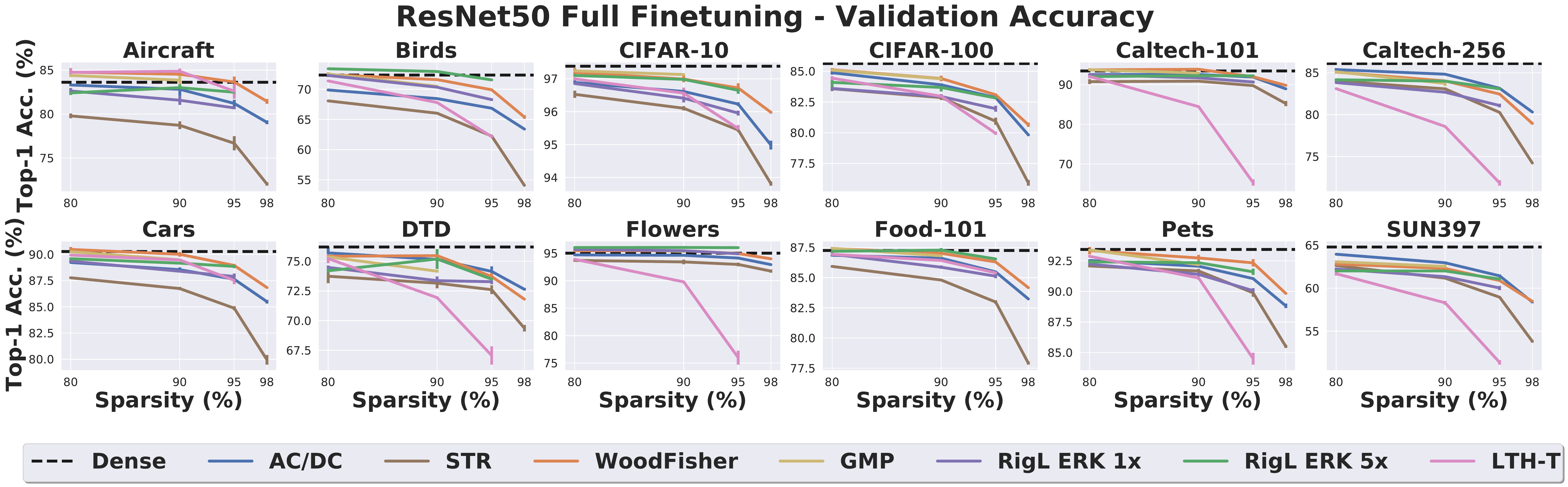}
    \caption{(ResNet50) Per-dataset downstream validation accuracy for transfer learining with \emph{full finetuning}.}
    \label{fig:individual_resnet_full}
\end{figure}

Additionally, we also validate our linear finetuning results by training with a different optimizer than SGD with momentum; namely, we use L-BFGS \cite{lbfgs} and $L_2$ regularization. We tested multiple values for the $L_2$ regularization strength and we report for each dataset and method the highest value for the test accuracy. The results of this experiment are presented in Table~\ref{table:rn50_linear_lbfgs}. Despite the differences in test accuracy between models trained with SGD or L-BFGS, we can observe a very similar trend related to the performance of sparse models over the dense baseline: namely, regularization pruning methods, such as AC/DC, STR or RigL, tend to be close to---or even outperform---the dense baseline, especially on fine-grained tasks (Aircraft, Cars).


\begin{table}[htb!]
\centering

\scalebox{0.9}{%
\begin{tabular}{ccccccccc}
\toprule
{Pruning Strategy} & {Dense} & {AC/DC} & {GMP} & {LTH-T} & {RigL ERK 1x} & {RigL ERK 5x} & {STR} & {WoodFisher} \\

\midrule
{80\% Sparsity} & {} & {} & {} & {} & {} & {} & {}\\
\midrule
Aircraft & 49.2 $ \pm $ 0.1 & \textbf{55.1 $ \pm $ 0.1} & 45.8 $ \pm $ 0.1 & 36.9 $ \pm $ 0.1 & 54.6 $ \pm $ 0.1 & \textbf{55.2 $ \pm $ 0.2} & 53.7 $ \pm $ 0.0 & 40.0 $ \pm $ 0.2 \\
Birds & 57.7 $ \pm $ 0.1 & \textbf{58.4 $ \pm $ 0.0} & 56.2 $ \pm $ 0.0 & 29.6 $ \pm $ 0.1 & 55.2 $ \pm $ 0.0 & 56.7 $ \pm $ 0.1 & 56.2 $ \pm $ 0.1 & 51.9 $ \pm $ 0.1 \\
CIFAR-10 & 91.2 $ \pm $ 0.0 & 90.9 $ \pm $ 0.0 & 89.7 $ \pm $ 0.0 & 83.4 $ \pm $ 0.1 & 89.7 $ \pm $ 0.1 & 90.0 $ \pm $ 0.1 & \textbf{91.4 $ \pm $ 0.0} & 89.6 $ \pm $ 0.0 \\
CIFAR-100 & \textbf{74.6 $ \pm $ 0.1} & \textbf{74.7 $ \pm $ 0.1} & 72.0 $ \pm $ 0.1 & 62.0 $ \pm $ 0.1 & 73.1 $ \pm $ 0.1 & 73.7 $ \pm $ 0.0 & \textbf{74.7 $ \pm $ 0.0} & 71.3 $ \pm $ 0.0 \\
Caltech-101 & 91.9 $ \pm $ 0.1 & \textbf{92.4 $ \pm $ 0.2} & 91.5 $ \pm $ 0.2 & 75.4 $ \pm $ 0.1 & 91.1 $ \pm $ 0.1 & 90.8 $ \pm $ 0.3 & 91.2 $ \pm $ 0.1 & 91.2 $ \pm $ 0.1 \\
Caltech-256 & \textbf{84.8 $ \pm $ 0.1} & \textbf{84.6 $ \pm $ 0.1} & 83.9 $ \pm $ 0.1 & 66.1 $ \pm $ 0.1 & 83.3 $ \pm $ 0.1 & \textbf{84.6 $ \pm $ 0.1} & 83.6 $ \pm $ 0.0 & 83.7 $ \pm $ 0.1 \\
Cars & 53.4 $ \pm $ 0.1 & 56.6 $ \pm $ 0.0 & 49.1 $ \pm $ 0.1 & 32.7 $ \pm $ 0.1 & 57.4 $ \pm $ 0.1 & \textbf{58.6 $ \pm $ 0.1} & 57.0 $ \pm $ 0.1 & 44.9 $ \pm $ 0.1 \\
DTD & 73.5 $ \pm $ 0.2 & \textbf{74.4 $ \pm $ 0.1} & 71.2 $ \pm $ 0.1 & 64.9 $ \pm $ 0.2 & 73.5 $ \pm $ 0.2 & 72.9 $ \pm $ 0.3 & \textbf{74.3 $ \pm $ 0.2} & 70.8 $ \pm $ 0.2 \\
Flowers & 91.6 $ \pm $ 0.1 & 92.7 $ \pm $ 0.1 & 90.9 $ \pm $ 0.1 & 85.6 $ \pm $ 0.1 & 92.2 $ \pm $ 0.1 & 92.3 $ \pm $ 0.1 & \textbf{93.0 $ \pm $ 0.0} & 87.6 $ \pm $ 0.1 \\
Food-101 & 73.2 $ \pm $ 0.0 & 73.8 $ \pm $ 0.0 & 70.5 $ \pm $ 0.0 & 61.9 $ \pm $ 0.0 & 73.3 $ \pm $ 0.0 & 72.5 $ \pm $ 0.1 & \textbf{73.9 $ \pm $ 0.0} & 68.5 $ \pm $ 0.0 \\
Pets & \textbf{92.6 $ \pm $ 0.1} & 92.3 $ \pm $ 0.1 & \textbf{92.5 $ \pm $ 0.1} & 79.4 $ \pm $ 0.1 & 91.9 $ \pm $ 0.1 & \textbf{92.5 $ \pm $ 0.2} & 91.7 $ \pm $ 0.0 & 92.2 $ \pm $ 0.1 \\
SUN397 & 60.1 $ \pm $ 0.0 & \textbf{60.4 $ \pm $ 0.0} & 58.1 $ \pm $ 0.0 & 47.4 $ \pm $ 0.0 & 59.1 $ \pm $ 0.1 & 59.9 $ \pm $ 0.0 & 60.3 $ \pm $ 0.0 & 57.8 $ \pm $ 0.1 \\

\midrule
{90\% Sparsity} & {} & {} & {} & {} & {} & {} & {}\\
\midrule
Aircraft & 49.2 $ \pm $ 0.1 & 55.5 $ \pm $ 0.1 & 48.7 $ \pm $ 0.1 & 16.5 $ \pm $ 0.2 & 54.1 $ \pm $ 0.1 & \textbf{56.6 $ \pm $ 0.1} & 52.9 $ \pm $ 0.1 & 44.0 $ \pm $ 0.2 \\
Birds & 57.7 $ \pm $ 0.1 & \textbf{58.7 $ \pm $ 0.0} & 55.4 $ \pm $ 0.1 & 11.4 $ \pm $ 0.1 & 53.3 $ \pm $ 0.0 & 57.2 $ \pm $ 0.1 & 55.2 $ \pm $ 0.1 & 52.7 $ \pm $ 0.1 \\
CIFAR-10 & \textbf{91.2 $ \pm $ 0.0} & 91.0 $ \pm $ 0.0 & 89.4 $ \pm $ 0.0 & 67.0 $ \pm $ 0.1 & 90.0 $ \pm $ 0.1 & 90.2 $ \pm $ 0.1 & 90.6 $ \pm $ 0.0 & 88.9 $ \pm $ 0.0 \\
CIFAR-100 & \textbf{74.6 $ \pm $ 0.1} & 74.3 $ \pm $ 0.0 & 71.5 $ \pm $ 0.0 & 42.2 $ \pm $ 0.1 & 72.8 $ \pm $ 0.1 & 73.4 $ \pm $ 0.1 & 73.7 $ \pm $ 0.1 & 70.5 $ \pm $ 0.0 \\
Caltech-101 & 91.9 $ \pm $ 0.1 & \textbf{92.5 $ \pm $ 0.1} & 91.6 $ \pm $ 0.1 & 49.0 $ \pm $ 0.6 & 90.6 $ \pm $ 0.3 & 91.4 $ \pm $ 0.4 & 90.9 $ \pm $ 0.1 & 91.3 $ \pm $ 0.1 \\
Caltech-256 & \textbf{84.8 $ \pm $ 0.1} & 84.5 $ \pm $ 0.0 & 82.9 $ \pm $ 0.0 & 42.0 $ \pm $ 0.1 & 81.9 $ \pm $ 0.0 & 84.5 $ \pm $ 0.1 & 82.6 $ \pm $ 0.0 & 83.0 $ \pm $ 0.1 \\
Cars & 53.4 $ \pm $ 0.1 & 56.0 $ \pm $ 0.1 & 50.2 $ \pm $ 0.0 & 15.4 $ \pm $ 0.1 & 55.5 $ \pm $ 0.1 & \textbf{60.5 $ \pm $ 0.1} & 54.8 $ \pm $ 0.1 & 46.7 $ \pm $ 0.0 \\
DTD & \textbf{73.5 $ \pm $ 0.2} & \textbf{73.7 $ \pm $ 0.2} & 72.4 $ \pm $ 0.2 & 54.7 $ \pm $ 0.1 & 72.6 $ \pm $ 0.3 & 72.7 $ \pm $ 0.2 & \textbf{73.8 $ \pm $ 0.1} & 71.0 $ \pm $ 0.2 \\
Flowers & 91.6 $ \pm $ 0.1 & 92.4 $ \pm $ 0.0 & 91.4 $ \pm $ 0.1 & 67.7 $ \pm $ 0.1 & 91.6 $ \pm $ 0.1 & 92.4 $ \pm $ 0.1 & \textbf{93.0 $ \pm $ 0.1} & 89.0 $ \pm $ 0.1 \\
Food-101 & 73.2 $ \pm $ 0.0 & \textbf{73.8 $ \pm $ 0.0} & 71.1 $ \pm $ 0.0 & 46.9 $ \pm $ 0.0 & 71.7 $ \pm $ 0.0 & 72.6 $ \pm $ 0.0 & 72.6 $ \pm $ 0.0 & 69.2 $ \pm $ 0.0 \\
Pets & \textbf{92.6 $ \pm $ 0.1} & 91.9 $ \pm $ 0.1 & 92.0 $ \pm $ 0.1 & 43.8 $ \pm $ 0.2 & 91.1 $ \pm $ 0.1 & 91.9 $ \pm $ 0.2 & 91.1 $ \pm $ 0.1 & 92.0 $ \pm $ 0.1 \\
SUN397 & \textbf{60.1 $ \pm $ 0.0} & 59.8 $ \pm $ 0.1 & 58.1 $ \pm $ 0.0 & 31.7 $ \pm $ 0.1 & 57.7 $ \pm $ 0.0 & 59.8 $ \pm $ 0.1 & 58.2 $ \pm $ 0.0 & 56.8 $ \pm $ 0.0 \\

\midrule
{95\% Sparsity} & {} & {} & {} & {} & {} & {} & {}\\
\midrule
Aircraft & 49.2 $ \pm $ 0.1 & 56.6 $ \pm $ 0.1 & {N \slash A} & 4.5 $ \pm $ 0.3 & 53.5 $ \pm $ 0.1 & \textbf{56.9 $ \pm $ 0.1} & 50.3 $ \pm $ 0.1 & 45.6 $ \pm $ 0.3 \\
Birds & \textbf{57.7 $ \pm $ 0.1} & \textbf{57.7 $ \pm $ 0.0} & {N \slash A} & 2.3 $ \pm $ 0.1 & 51.9 $ \pm $ 0.1 & 55.9 $ \pm $ 0.0 & 52.1 $ \pm $ 0.1 & 51.8 $ \pm $ 0.1 \\
CIFAR-10 & \textbf{91.2 $ \pm $ 0.0} & 90.5 $ \pm $ 0.0 & {N \slash A} & 39.9 $ \pm $ 0.2 & 89.4 $ \pm $ 0.0 & 89.8 $ \pm $ 0.1 & 89.1 $ \pm $ 0.0 & 88.6 $ \pm $ 0.0 \\
CIFAR-100 & \textbf{74.6 $ \pm $ 0.1} & 73.4 $ \pm $ 0.0 & {N \slash A} & 13.5 $ \pm $ 0.2 & 71.5 $ \pm $ 0.1 & 72.4 $ \pm $ 0.1 & 71.7 $ \pm $ 0.0 & 69.7 $ \pm $ 0.0 \\
Caltech-101 & \textbf{91.9 $ \pm $ 0.1} & 91.6 $ \pm $ 0.1 & {N \slash A} & 20.1 $ \pm $ 0.5 & 89.0 $ \pm $ 0.1 & 91.4 $ \pm $ 0.1 & 90.0 $ \pm $ 0.2 & 91.0 $ \pm $ 0.2 \\
Caltech-256 & \textbf{84.8 $ \pm $ 0.1} & 82.8 $ \pm $ 0.1 & {N \slash A} & 12.4 $ \pm $ 0.3 & 80.1 $ \pm $ 0.1 & 83.5 $ \pm $ 0.1 & 80.2 $ \pm $ 0.1 & 81.2 $ \pm $ 0.1 \\
Cars & 53.4 $ \pm $ 0.1 & \textbf{56.9 $ \pm $ 0.1} & {N \slash A} & 3.9 $ \pm $ 0.1 & 52.9 $ \pm $ 0.0 & \textbf{57.0 $ \pm $ 0.1} & 50.5 $ \pm $ 0.1 & 45.5 $ \pm $ 0.0 \\
DTD & \textbf{73.5 $ \pm $ 0.2} & 72.7 $ \pm $ 0.1 & {N \slash A} & 27.4 $ \pm $ 0.2 & 71.9 $ \pm $ 0.1 & 72.9 $ \pm $ 0.2 & 72.1 $ \pm $ 0.2 & 70.4 $ \pm $ 0.1 \\
Flowers & 91.6 $ \pm $ 0.1 & \textbf{93.0 $ \pm $ 0.1} & {N \slash A} & 27.8 $ \pm $ 0.6 & 91.0 $ \pm $ 0.1 & 92.4 $ \pm $ 0.1 & 91.9 $ \pm $ 0.1 & 89.6 $ \pm $ 0.0 \\
Food-101 & \textbf{73.2 $ \pm $ 0.0} & \textbf{73.2 $ \pm $ 0.0} & {N \slash A} & 15.0 $ \pm $ 0.1 & 70.6 $ \pm $ 0.1 & 71.9 $ \pm $ 0.0 & 70.7 $ \pm $ 0.0 & 68.2 $ \pm $ 0.0 \\
Pets & \textbf{92.6 $ \pm $ 0.1} & 91.0 $ \pm $ 0.2 & {N \slash A} & 15.9 $ \pm $ 0.2 & 90.1 $ \pm $ 0.1 & 91.1 $ \pm $ 0.1 & 89.8 $ \pm $ 0.1 & 91.4 $ \pm $ 0.0 \\
SUN397 & \textbf{60.1 $ \pm $ 0.0} & 58.2 $ \pm $ 0.0 & {N \slash A} & 8.4 $ \pm $ 0.2 & 55.9 $ \pm $ 0.1 & 58.3 $ \pm $ 0.1 & 56.3 $ \pm $ 0.0 & 55.1 $ \pm $ 0.1 \\

\midrule
{98\% Sparsity} & {} & {} & {} & {} & {} & {} \\
\midrule
Aircraft & 49.2 $ \pm $ 0.1 & \textbf{54.8 $ \pm $ 0.1} & {N \slash A} & {N \slash A} & {N \slash A} & {N \slash A} & 48.0 $ \pm $ 0.1 & 45.0 $ \pm $ 0.1 \\
Birds & \textbf{57.7 $ \pm $ 0.1} & 54.5 $ \pm $ 0.0 & {N \slash A} & {N \slash A} & {N \slash A} & {N \slash A} & 43.7 $ \pm $ 0.0 & 48.1 $ \pm $ 0.1 \\
CIFAR-10 & \textbf{91.2 $ \pm $ 0.0} & 89.2 $ \pm $ 0.0 & {N \slash A} & {N \slash A} & {N \slash A} & {N \slash A} & 86.5 $ \pm $ 0.0 & 86.6 $ \pm $ 0.0 \\
CIFAR-100 & \textbf{74.6 $ \pm $ 0.1} & 71.6 $ \pm $ 0.0 & {N \slash A} & {N \slash A} & {N \slash A} & {N \slash A} & 67.4 $ \pm $ 0.0 & 67.8 $ \pm $ 0.0 \\
Caltech-101 & \textbf{91.9 $ \pm $ 0.1} & 89.0 $ \pm $ 0.1 & {N \slash A} & {N \slash A} & {N \slash A} & {N \slash A} & 86.3 $ \pm $ 0.1 & 88.5 $ \pm $ 0.1 \\
Caltech-256 & \textbf{84.8 $ \pm $ 0.1} & 79.8 $ \pm $ 0.0 & {N \slash A} & {N \slash A} & {N \slash A} & {N \slash A} & 73.4 $ \pm $ 0.1 & 77.1 $ \pm $ 0.0 \\
Cars & \textbf{53.4 $ \pm $ 0.1} & 52.1 $ \pm $ 0.0 & {N \slash A} & {N \slash A} & {N \slash A} & {N \slash A} & 44.4 $ \pm $ 0.1 & 42.2 $ \pm $ 0.0 \\
DTD & \textbf{73.5 $ \pm $ 0.2} & 71.6 $ \pm $ 0.1 & {N \slash A} & {N \slash A} & {N \slash A} & {N \slash A} & 68.4 $ \pm $ 0.2 & 68.3 $ \pm $ 0.1 \\
Flowers & 91.6 $ \pm $ 0.1 & \textbf{92.3 $ \pm $ 0.1} & {N \slash A} & {N \slash A} & {N \slash A} & {N \slash A} & 90.8 $ \pm $ 0.1 & 89.5 $ \pm $ 0.1 \\
Food-101 & \textbf{73.2 $ \pm $ 0.0} & 70.8 $ \pm $ 0.0 & {N \slash A} & {N \slash A} & {N \slash A} & {N \slash A} & 65.3 $ \pm $ 0.0 & 66.5 $ \pm $ 0.0 \\
Pets & \textbf{92.6 $ \pm $ 0.1} & 89.2 $ \pm $ 0.1 & {N \slash A} & {N \slash A} & {N \slash A} & {N \slash A} & 85.5 $ \pm $ 0.1 & 88.7 $ \pm $ 0.1 \\
SUN397 & \textbf{60.1 $ \pm $ 0.0} & 55.1 $ \pm $ 0.0 & {N \slash A} & {N \slash A} & {N \slash A} & {N \slash A} & 50.9 $ \pm $ 0.0 & 52.4 $ \pm $ 0.0 \\
\bottomrule
\end{tabular}
}
\caption{Transfer accuracy for sparse ResNet50 transfer with \emph{linear finetuning}.}
\label{table:rn50_linear_all}
\end{table}


\subsection{Full Finetuning Results}
\label{appendix:rn50_full}

Similarly to linear finetuning, we further provide complete results for full finetuning from sparse models. We present individual results per downstream task and pruning method, at different sparsity levels, in Figure~\ref{fig:individual_resnet_full} and Table~\ref{table:rn50_full_all}; we report for each the average and standard deviation across 3 different trials. The results further support our conclusions from Section~\ref{subsec:full-finetuning}; namely, downstream task accuracy is correlated with the backbone sparsity, and progressive sparsification methods (GMP, WoodFisher) generally perform better than regularization methods.  

\begin{table}[htb!]
\centering
\scalebox{0.9}{
\begin{tabular}{ccccccccc}
\toprule
{Pruning Strategy} & {Dense} & {AC/DC} & {GMP} & {LTH-T} & {RigL ERK 1x} & {RigL ERK 5x} & {STR} & {WoodFisher} \\
\midrule
{80\% Sparsity} & {} & {} & {} & {} & {} & {} & {} & {}\\
\midrule

Aircraft & 83.6 $ \pm $ 0.4 & 83.3 $ \pm $ 0.1 & \textbf{84.4 $ \pm $ 0.2} & \textbf{84.7 $ \pm $ 0.5} & 82.6 $ \pm $ 0.3 & 82.4 $ \pm $ 0.2 & 79.8 $ \pm $ 0.3 & \textbf{84.8 $ \pm $ 0.2} \\

Birds & 72.4 $ \pm $ 0.3 & 69.9 $ \pm $ 0.2 & 72.5 $ \pm $ 0.2 & 71.4 $ \pm $ 0.1 & 72.3 $ \pm $ 0.3 & \textbf{73.4 $ \pm $ 0.1} & 68.1 $ \pm $ 0.1 & 72.4 $ \pm $ 0.4 \\

CIFAR-10 & \textbf{97.4 $ \pm $ 0.0} & 96.9 $ \pm $ 0.1 & 97.2 $ \pm $ 0.0 & 97.0 $ \pm $ 0.0 & 96.9 $ \pm $ 0.0 & 97.1 $ \pm $ 0.0 & 96.5 $ \pm $ 0.1 & 97.2 $ \pm $ 0.1 \\

CIFAR-100 & \textbf{85.6 $ \pm $ 0.2} & 84.9 $ \pm $ 0.2 & 85.1 $ \pm $ 0.0 & 84.4 $ \pm $ 0.2 & 83.6 $ \pm $ 0.2 & 84.1 $ \pm $ 0.4 & 83.6 $ \pm $ 0.2 & 85.1 $ \pm $ 0.1 \\
Caltech-101 & \textbf{93.5 $ \pm $ 0.1} & 92.5 $ \pm $ 0.2 & \textbf{93.7 $ \pm $ 0.5} & 92.1 $ \pm $ 0.5 & 92.5 $ \pm $ 0.1 & 92.0 $ \pm $ 0.3 & 90.7 $ \pm $ 0.6 & \textbf{93.7 $ \pm $ 0.1} \\

Caltech-256 & \textbf{86.1 $ \pm $ 0.1} & 85.4 $ \pm $ 0.2 & 85.1 $ \pm $ 0.2 & 83.1 $ \pm $ 0.1 & 83.8 $ \pm $ 0.1 & 84.2 $ \pm $ 0.2 & 84.0 $ \pm $ 0.1 & 85.1 $ \pm $ 0.1 \\

Cars & \textbf{90.3 $ \pm $ 0.2} & 89.2 $ \pm $ 0.1 & \textbf{90.3 $ \pm $ 0.1} & 89.9 $ \pm $ 0.0 & 89.4 $ \pm $ 0.1 & 89.6 $ \pm $ 0.1 & 87.8 $ \pm $ 0.1 & \textbf{90.5 $ \pm $ 0.2} \\

DTD & \textbf{76.2 $ \pm $ 0.3} & 75.7 $ \pm $ 0.5 & 75.4 $ \pm $ 0.1 & 75.2 $ \pm $ 0.4 & 74.5 $ \pm $ 0.2 & 74.2 $ \pm $ 0.2 & 73.7 $ \pm $ 0.6 & 75.4 $ \pm $ 0.3 \\

Flowers & 95.0 $ \pm $ 0.1 & 94.7 $ \pm $ 0.2 & 95.9 $ \pm $ 0.2 & 93.9 $ \pm $ 0.2 & 95.7 $ \pm $ 0.2 & \textbf{96.1 $ \pm $ 0.1} & 93.7 $ \pm $ 0.2 & 95.5 $ \pm $ 0.2 \\

Food-101 & \textbf{87.3 $ \pm $ 0.1} & 86.9 $ \pm $ 0.1 & \textbf{87.4 $ \pm $ 0.1} & 86.9 $ \pm $ 0.1 & 86.9 $ \pm $ 0.1 & \textbf{87.2 $ \pm $ 0.1} & 85.9 $ \pm $ 0.1 & \textbf{87.4 $ \pm $ 0.1} \\

Pets & \textbf{93.4 $ \pm $ 0.1} & 92.5 $ \pm $ 0.0 & \textbf{93.4 $ \pm $ 0.1} & 92.9 $ \pm $ 0.1 & 92.2 $ \pm $ 0.1 & 92.4 $ \pm $ 0.1 & 92.1 $ \pm $ 0.1 & \textbf{93.3 $ \pm $ 0.3} \\

SUN397 & \textbf{64.8 $ \pm $ 0.0} & 64.0 $ \pm $ 0.0 & 63.1 $ \pm $ 0.1 & 61.7 $ \pm $ 0.2 & 62.2 $ \pm $ 0.2 & 62.0 $ \pm $ 0.3 & 62.6 $ \pm $ 0.1 & 62.8 $ \pm $ 0.1 \\

\midrule
{90\% Sparsity} & {} & {} & {} & {} & {} & {} & {} & {} \\
\midrule

Aircraft & 83.6 $ \pm $ 0.4 & 82.8 $ \pm $ 1.0 & \textbf{83.9 $ \pm $ 0.7} & \textbf{84.9 $ \pm $ 0.3} & 81.6 $ \pm $ 0.5 & 83.0 $ \pm $ 0.4 & 78.7 $ \pm $ 0.4 & \textbf{84.5 $ \pm $ 0.4} \\
Birds & \textbf{72.4 $ \pm $ 0.3} & 68.5 $ \pm $ 0.1 & 70.5 $ \pm $ 0.1 & 67.8 $ \pm $ 0.2 & 70.3 $ \pm $ 0.0 & \textbf{72.9 $ \pm $ 0.2} & 66.0 $ \pm $ 0.2 & 71.6 $ \pm $ 0.2 \\
CIFAR-10 & \textbf{97.4 $ \pm $ 0.0} & 96.6 $ \pm $ 0.1 & 97.1 $ \pm $ 0.0 & 96.6 $ \pm $ 0.2 & 96.4 $ \pm $ 0.1 & 97.0 $ \pm $ 0.1 & 96.1 $ \pm $ 0.1 & 97.0 $ \pm $ 0.1 \\
CIFAR-100 & \textbf{85.6 $ \pm $ 0.2} & 83.9 $ \pm $ 0.1 & 84.4 $ \pm $ 0.0 & 83.0 $ \pm $ 0.1 & 83.0 $ \pm $ 0.2 & 83.7 $ \pm $ 0.3 & 82.9 $ \pm $ 0.2 & 84.4 $ \pm $ 0.2 \\
Caltech-101 & \textbf{93.5 $ \pm $ 0.1} & 92.6 $ \pm $ 0.2 & 92.9 $ \pm $ 0.2 & 84.5 $ \pm $ 0.3 & 91.7 $ \pm $ 0.3 & 92.3 $ \pm $ 0.4 & 90.9 $ \pm $ 0.3 & \textbf{93.9 $ \pm $ 0.3} \\
Caltech-256 & \textbf{86.1 $ \pm $ 0.1} & 84.8 $ \pm $ 0.1 & 83.7 $ \pm $ 0.3 & 78.6 $ \pm $ 0.1 & 82.7 $ \pm $ 0.2 & 84.0 $ \pm $ 0.1 & 83.1 $ \pm $ 0.2 & 84.0 $ \pm $ 0.1 \\
Cars & \textbf{90.3 $ \pm $ 0.2} & 88.5 $ \pm $ 0.2 & 89.5 $ \pm $ 0.0 & 89.5 $ \pm $ 0.1 & 88.4 $ \pm $ 0.1 & 89.2 $ \pm $ 0.1 & 86.7 $ \pm $ 0.2 & \textbf{90.0 $ \pm $ 0.2} \\
DTD & \textbf{76.2 $ \pm $ 0.3} & 75.2 $ \pm $ 0.1 & 74.2 $ \pm $ 0.1 & 71.9 $ \pm $ 0.1 & 73.4 $ \pm $ 0.4 & 75.2 $ \pm $ 0.8 & 73.2 $ \pm $ 0.4 & 75.5 $ \pm $ 0.4 \\
Flowers & 95.0 $ \pm $ 0.1 & 94.6 $ \pm $ 0.1 & 95.3 $ \pm $ 0.1 & 89.8 $ \pm $ 0.2 & 95.5 $ \pm $ 0.1 & \textbf{96.1 $ \pm $ 0.1} & 93.4 $ \pm $ 0.4 & 95.5 $ \pm $ 0.3 \\
Food-101 & \textbf{87.3 $ \pm $ 0.1} & 86.6 $ \pm $ 0.1 & 86.8 $ \pm $ 0.1 & 86.4 $ \pm $ 0.1 & 85.9 $ \pm $ 0.1 & \textbf{87.3 $ \pm $ 0.2} & 84.8 $ \pm $ 0.0 & \textbf{87.0 $ \pm $ 0.1} \\
Pets & \textbf{93.4 $ \pm $ 0.1} & 92.1 $ \pm $ 0.1 & 92.2 $ \pm $ 0.1 & 91.1 $ \pm $ 0.2 & 91.4 $ \pm $ 0.2 & 92.3 $ \pm $ 0.1 & 91.7 $ \pm $ 0.2 & 92.7 $ \pm $ 0.3 \\
SUN397 & \textbf{64.8 $ \pm $ 0.0} & 63.0 $ \pm $ 0.0 & 62.5 $ \pm $ 0.2 & 58.3 $ \pm $ 0.2 & 61.3 $ \pm $ 0.1 & 62.0 $ \pm $ 0.2 & 61.2 $ \pm $ 0.0 & 62.3 $ \pm $ 0.1 \\

\midrule
{95\% Sparsity} & {} & {} & {} & {} & {} & {} & {} \\
\midrule
Aircraft & \textbf{83.6 $ \pm $ 0.4} & 81.2 $ \pm $ 0.4 & {N \slash A} & 82.6 $ \pm $ 0.8 & 80.7 $ \pm $ 0.1 & 82.5 $ \pm $ 0.4 & 76.7 $ \pm $ 0.8 & \textbf{83.6 $ \pm $ 0.6} \\
Birds & \textbf{72.4 $ \pm $ 0.3} & 66.9 $ \pm $ 0.1 & {N \slash A} & 62.2 $ \pm $ 0.1 & 68.3 $ \pm $ 0.2 & 71.6 $ \pm $ 0.1 & 62.3 $ \pm $ 0.1 & 69.9 $ \pm $ 0.1 \\
CIFAR-10 & \textbf{97.4 $ \pm $ 0.0} & 96.2 $ \pm $ 0.1 & {N \slash A} & 95.5 $ \pm $ 0.1 & 96.0 $ \pm $ 0.1 & 96.6 $ \pm $ 0.1 & 95.4 $ \pm $ 0.1 & 96.7 $ \pm $ 0.1 \\
CIFAR-100 & \textbf{85.6 $ \pm $ 0.2} & 82.9 $ \pm $ 0.1 & {N \slash A} & 80.0 $ \pm $ 0.1 & 82.0 $ \pm $ 0.2 & 82.8 $ \pm $ 0.0 & 80.9 $ \pm $ 0.3 & 83.1 $ \pm $ 0.1 \\
Caltech-101 & \textbf{93.5 $ \pm $ 0.1} & 91.9 $ \pm $ 0.2 & {N \slash A} & 65.3 $ \pm $ 0.8 & 90.7 $ \pm $ 0.4 & 92.2 $ \pm $ 0.3 & 89.8 $ \pm $ 0.1 & 92.0 $ \pm $ 0.3 \\
Caltech-256 & \textbf{86.1 $ \pm $ 0.1} & 83.1 $ \pm $ 0.0 & {N \slash A} & 71.8 $ \pm $ 0.3 & 81.1 $ \pm $ 0.2 & 83.1 $ \pm $ 0.2 & 80.3 $ \pm $ 0.0 & 82.4 $ \pm $ 0.1 \\
Cars & \textbf{90.3 $ \pm $ 0.2} & 87.6 $ \pm $ 0.1 & {N \slash A} & 87.5 $ \pm $ 0.4 & 87.9 $ \pm $ 0.3 & 88.9 $ \pm $ 0.2 & 84.9 $ \pm $ 0.2 & 88.9 $ \pm $ 0.2 \\
DTD & \textbf{76.2 $ \pm $ 0.3} & 74.1 $ \pm $ 0.4 & {N \slash A} & 67.1 $ \pm $ 0.8 & 73.3 $ \pm $ 0.2 & 73.5 $ \pm $ 0.2 & 72.6 $ \pm $ 0.4 & 73.7 $ \pm $ 0.3 \\
Flowers & 95.0 $ \pm $ 0.1 & 94.1 $ \pm $ 0.3 & {N \slash A} & 76.0 $ \pm $ 1.3 & 94.9 $ \pm $ 0.3 & \textbf{96.0 $ \pm $ 0.0} & 93.0 $ \pm $ 0.3 & 95.0 $ \pm $ 0.3 \\
Food-101 & \textbf{87.3 $ \pm $ 0.1} & 85.5 $ \pm $ 0.0 & {N \slash A} & 85.4 $ \pm $ 0.1 & 85.1 $ \pm $ 0.2 & 86.6 $ \pm $ 0.0 & 83.0 $ \pm $ 0.1 & 86.3 $ \pm $ 0.1 \\
Pets & \textbf{93.4 $ \pm $ 0.1} & 91.0 $ \pm $ 0.1 & {N \slash A} & 84.5 $ \pm $ 0.5 & 90.1 $ \pm $ 0.2 & 91.6 $ \pm $ 0.3 & 89.9 $ \pm $ 0.3 & 92.3 $ \pm $ 0.3 \\
SUN397 & \textbf{64.8 $ \pm $ 0.0} & 61.4 $ \pm $ 0.2 & {N \slash A} & 51.4 $ \pm $ 0.3 & 60.0 $ \pm $ 0.3 & 61.1 $ \pm $ 0.2 & 59.0 $ \pm $ 0.1 & 60.9 $ \pm $ 0.1 \\

\midrule
{98\% Sparsity} & {} & {} & {} & {} & {} & {} & {} & {} \\
\midrule
Aircraft & \textbf{83.6 $ \pm $ 0.4} & 79.1 $ \pm $ 0.2 & {N \slash A} & {N \slash A} & {N \slash A} & {N \slash A} & 72.0 $ \pm $ 0.2 & 81.4 $ \pm $ 0.3 \\
Birds & \textbf{72.4 $ \pm $ 0.3} & 63.4 $ \pm $ 0.1 & {N \slash A} & {N \slash A} & {N \slash A} & {N \slash A} & 54.1 $ \pm $ 0.1 & 65.4 $ \pm $ 0.3 \\
CIFAR-10 & \textbf{97.4 $ \pm $ 0.0} & 95.0 $ \pm $ 0.1 & {N \slash A} & {N \slash A} & {N \slash A} & {N \slash A} & 93.8 $ \pm $ 0.1 & 96.0 $ \pm $ 0.0 \\
CIFAR-100 & \textbf{85.6 $ \pm $ 0.2} & 79.8 $ \pm $ 0.1 & {N \slash A} & {N \slash A} & {N \slash A} & {N \slash A} & 75.9 $ \pm $ 0.2 & 80.7 $ \pm $ 0.2 \\
Caltech-101 & \textbf{93.5 $ \pm $ 0.1} & 88.9 $ \pm $ 0.1 & {N \slash A} & {N \slash A} & {N \slash A} & {N \slash A} & 85.2 $ \pm $ 0.6 & 89.8 $ \pm $ 0.3 \\
Caltech-256 & \textbf{86.1 $ \pm $ 0.1} & 80.3 $ \pm $ 0.1 & {N \slash A} & {N \slash A} & {N \slash A} & {N \slash A} & 74.2 $ \pm $ 0.0 & 78.9 $ \pm $ 0.1 \\
Cars & \textbf{90.3 $ \pm $ 0.2} & 85.5 $ \pm $ 0.2 & {N \slash A} & {N \slash A} & {N \slash A} & {N \slash A} & 79.9 $ \pm $ 0.5 & 86.8 $ \pm $ 0.1 \\
DTD & \textbf{76.2 $ \pm $ 0.3} & 72.6 $ \pm $ 0.1 & {N \slash A} & {N \slash A} & {N \slash A} & {N \slash A} & 69.4 $ \pm $ 0.3 & 71.8 $ \pm $ 0.1 \\
Flowers & \textbf{95.0 $ \pm $ 0.1} & 92.9 $ \pm $ 0.1 & {N \slash A} & {N \slash A} & {N \slash A} & {N \slash A} & 91.8 $ \pm $ 0.3 & 94.0 $ \pm $ 0.2 \\
Food-101 & \textbf{87.3 $ \pm $ 0.1} & 83.2 $ \pm $ 0.0  & {N \slash A} & {N \slash A} & {N \slash A} & {N \slash A} & 77.9 $ \pm $ 0.1 & 84.2 $ \pm $ 0.1 \\
Pets & \textbf{93.4 $ \pm $ 0.1} & 88.8 $ \pm $ 0.2 & {N \slash A} & {N \slash A} & {N \slash A} & {N \slash A} & 85.5 $ \pm $ 0.1 & 89.8 $ \pm $ 0.1 \\
SUN397 & \textbf{64.8 $ \pm $ 0.0} & 58.4 $ \pm $ 0.1 & {N \slash A} & {N \slash A} & {N \slash A} & {N \slash A} & 53.8 $ \pm $ 0.2 & 58.5 $ \pm $ 0.1 \\
\bottomrule
\end{tabular}

}
\caption{Transfer accuracy for sparse ResNet50 transfer with \emph{full finetuning}.}
\label{table:rn50_full_all}
\end{table}


\begin{table}[htb!]
\centering

\scalebox{0.9}{%
\begin{tabular}{cccccccc}
\toprule
{Pruning Strategy} & {Dense} & {AC/DC} & {GMP} & {RigL ERK 1x} & {RigL ERK 5x} & {STR} & {WoodFisher} \\
\midrule 
{80\% Sparsity} & {} & {} & {} & {} & {} & {} & {} \\
\midrule 
Aircraft & 50.3 & \textbf{56.7} & 46.9 & 55.4 & 55.6 & 54.6 & 43.1 \\
Birds & 56.7 & \textbf{57.7} & 54.6 & 55.1 & 56.2 & 55.8 & 50.7 \\
Caltech-101 & 91.8 & \textbf{92.0} & 91.2 & 91.5 &  91.2 & 91.4 & 91.3 \\
Caltech-256 & 84.3 & \textbf{84.6} & 83.2 & 83.3 & \textbf{84.6} & 83.3 & 83.0 \\
Cars & 56.2 & 59.5 & 50.1 & 58.9 & \textbf{60.4} & 60.0 & 46.5 \\
CIFAR-10 & 88.5 & 88.3 & 87.5 & 86.9 & 88.1 & \textbf{88.9} & 86.3 \\
CIFAR-100 & 72.3 & 72.4 & 69.1 & 70.7 & 71.8 & \textbf{72.9} & 68.1 \\
DTD & 73.2 & 72.8 & 69.9 & 72.9 & 73.1 & \textbf{73.3} & 70.0 \\
Flowers & 92.9 & \textbf{93.9} & 92.0 & 93.3 & 93.3 & \textbf{93.9} & 89.0 \\
Food-101 & 67.7 & \textbf{68.6} & 65.3 & 67.2 & 68.1 & 68.1 & 62.8 \\
Pets & \textbf{92.5} & 91.9 & 92.2 & 91.3 & 92.2 & 91.5 & 91.4 \\
SUN397 & 58.5 & 59.3 & 56.4 & 58.0 & \textbf{59.4} & \textbf{59.4} & 55.8 \\

\midrule 
90\% Sparsity & & & & & & & \\
\midrule
Aircraft & 50.3 & 56.7 & 49.6 & 55.3 & \textbf{57.4} & 54.6 & 45.0 \\
Birds & 56.7 & \textbf{57.7} & 54.3 & 52.8 & 56.9 & 54.7 & 51.5 \\
Caltech-101 & 91.8 & \textbf{92.3} & 91.0 & 90.5 & 91.5 & 90.4 & 91.2 \\
Caltech-256 & \textbf{84.3} & 84.1 & 82.6 & 81.7 & \textbf{84.7} & 82.3 & 82.5 \\
Cars & 56.2 & 59.0 & 52.4 & 57.3 & \textbf{62.0} & 57.8 & 48.4 \\
CIFAR-10 & \textbf{88.5} & \textbf{88.5} & 86.7 & 87.1 & 87.5 & 87.4 & 86.2 \\
CIFAR-100 & \textbf{72.3} & 71.6 & 68.9 & 70.1 & 72.0 & 72.0 & 67.6 \\
DTD & \textbf{73.2} & 72.8 & 71.8 & 71.5 & 71.6 & 72.2 & 69.3 \\
Flowers & 92.9 & 93.4 & 92.7 & 92.6 & 93.3 & \textbf{94.1} & 90.2 \\
Food-101 & \textbf{67.7} & \textbf{67.7} & 65.9 & 65.0 & 67.5 & 67.3 & 63.6 \\
Pets & \textbf{92.5} & 91.6 & 91.8 & 91.3 & 91.5 & 90.5 & 91.1 \\
SUN397 & 58.5 & 58.2 & 56.3 & 56.9 & \textbf{59.0} & 57.2 & 54.6 \\

\midrule 
95\% Sparsity & & & & & & & \\
\midrule 

Aircraft & 50.3 & 57.2 & {N \slash A} & 54.3 & \textbf{57.4} & 51.5 & 45.7 \\
Birds & \textbf{56.7} & 56.4 & {N \slash A} & 50.8 & 55.5 & 51.1 & 49.9 \\
Caltech-101 & \textbf{91.8} & 91.6 & {N \slash A} & 89.4 & 91.7 & 89.9 & 90.6 \\
Caltech-256 & \textbf{84.3} & 82.4 & {N \slash A} & 80.1 & 83.5 & 80.0 & 80.8 \\
Cars & 56.2 & \textbf{59.4} & {N \slash A} & 55.1 & 58.8 & 53.0 & 46.8 \\
CIFAR-10 & \textbf{88.5} & 87.9 & {N \slash A} & 86.7 & 86.9 & 86.4 & 86.3 \\
CIFAR-100 & \textbf{72.3} & 69.6 & {N \slash A} & 68.8 & 70.0 & 69.6 & 66.4 \\
DTD & \textbf{73.2} & 71.3 & {N \slash A} & 71.1 & 72.8 & 70.3 & 70.1 \\
Flowers & 92.9 & \textbf{94.2} & {N \slash A} & 92.3 & 93.5 & 93.1 & 90.8 \\
Food-101 & \textbf{67.7} & 66.6 & {N \slash A} & 63.6 & 66.1 & 64.8 & 63.0 \\
Pets & \textbf{92.5} & 90.4 & {N \slash A} & 89.7 & 90.9 & 89.2 & 90.5 \\
SUN397 & \textbf{58.5} & 56.8 & {N \slash A} & 54.9 & 57.7 & 54.8 & 52.7 \\

\bottomrule 
\end{tabular}}
\caption{Validation accuracy for sparse ResNet50 transfer with linear finetuning using the L-BFGS optimizer}
\label{table:rn50_linear_lbfgs}
\end{table}

\clearpage

\section{Results for Extended Training Schedule}
\label{appendix:extended-training}

\begin{table}[h]
\centering
\scalebox{0.7}{%
\begin{tabular}{@{}ccc@{}}
\toprule
 Sparsity  & 
Method & Imagenet Validation Accuracy \\
\toprule
$0\%$ & Dense & 76.8\% \\
 & \textbf{Dense 2x} & \textbf{77.1\% }\\

\midrule
$80\%$ & AC/DC & 76.2\%   \\
& \textbf{AC/DC 3x} & \textbf{77.5\% }\\
& RigL ERK 1x & 74.8\% \\
& RigL ERK 5x & 75.8\% \\

\midrule
$90\%$ & AC/DC & 75.2\%\\
& AC/DC 3x & 76.8\% \\
& \textbf{AC/DC 5x} & \textbf{77.2\%} \\
& RigL ERK 1x & 73.2\% \\
& RigL ERK 5x & 75.7\% \\

\midrule
$95\%$ & AC/DC & 73.1\% \\ 
& \textbf{AC/DC 3x} & \textbf{75.3\%} \\
& RigL ERK 1x & 70.1\% \\
& RigL ERK 5x & 74.0\%\\
\bottomrule
\end{tabular}
}
\caption{Accuracy of models with extended training time, evaluated on different ImageNet validation sets.} 
\label{table:extended_sparse_eval}
\end{table}

\begin{figure}[h]
    \centering
    \includegraphics[width=0.9\textwidth]{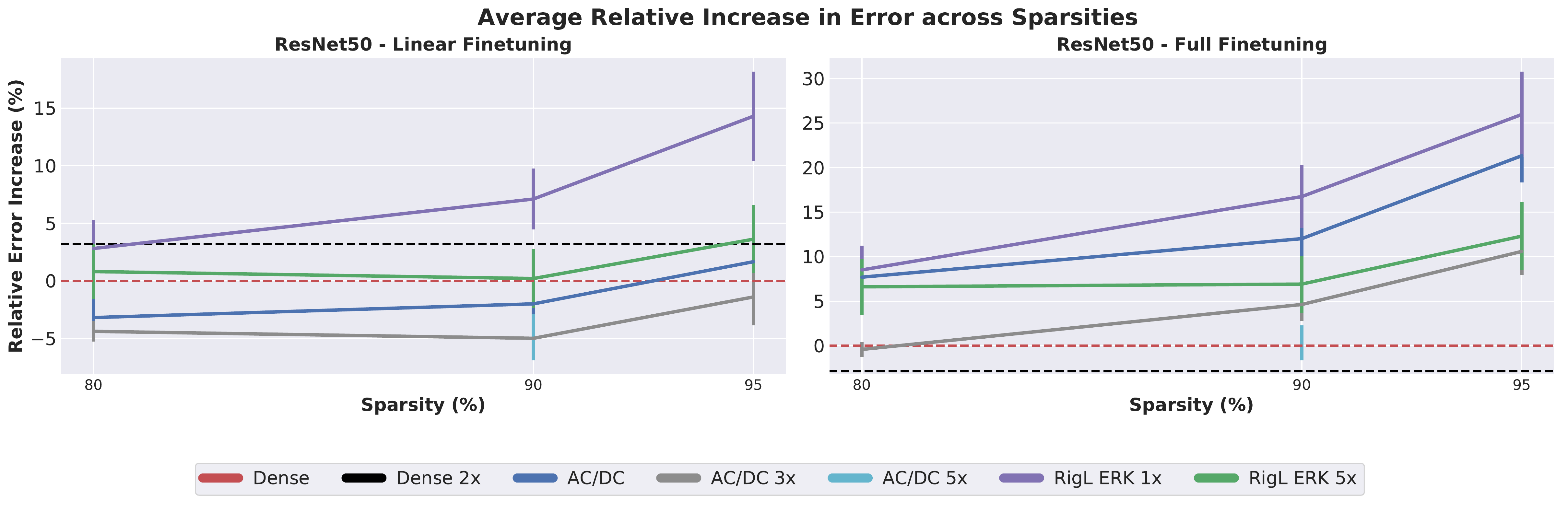}
    \caption{(ResNet50) Overall performance for extended-time runs of AC/DC and RigL. The AC/DC 5x was only done for 90\% sparsity; its performance is almost the same as AC/DC 3x for linear finetuning and as dense for full finetuning.}
    \label{fig:individual_resnet_longruns}
\end{figure}

Noting the improved performance of models compressed RigL ERK when running the sparse upstream training algorithm for 5x the number of epochs, we repeated the experiment with AC/DC, compressing ResNet50 models to 80\%, 90\%, and 95\% for 3x the training time (300 epochs) and to 90\% for 5x the training time (500 epochs). The latter experiment was only done at one compression level due to the high computational costs of upstream training. We also trained a dense model with 2x training time (200 epochs). We observe that extended training increases performance on ImageNet in all cases (see Table \ref{table:extended_sparse_eval}), especially for sparse models. In particular, extending training 3x raises the ImageNet validation accuracy by 1.3\% at 80\% sparsity (from 76.2\% to 77.5\%), by 1.5\% at 90\% (from 75.2\% to 76.7\%), and by 2.2\% at 95\% sparsity (from 73.1\% to 75.3\%). Training for 5x epochs at 90\% sparsity recovers the dense 2x validation accuracy. We see a similar trend with RigL ERK 1x/5x results. Thus, for regularization-based methods, extended training can substantially improve validation performance.

The same behavior extends to transfer learning. The complete results of the extended-training transfer experiments are presented in tables \ref{table:rn50_linear_longruns} for linear downstream finetuning and  \ref{table:rn50_full_longruns} for full downstream finetuning. We also summarize the results in figure \ref{fig:individual_resnet_longruns}, relying on the same average relative increase in error metric that we define in Section \ref{sec:techniques}. We observe that, while extended dense training overall gives an additional benefit for full finetuning balanced by a commensurate loss for linear finetuning, extended training of compressed models gives a performance improvement in both scenarios, for both RigL and AC/DC. Notably, 3x upstream training with AC/DC allows 80\% sparse models to perform as well as dense 1x models when doing full downstream finetuning, and 5x upstream training with AC/DC allows even 90\% sparse models to do so. These results suggest that for regularization-based methods, the extra investment in upstream training can result in upstream models that transfer very well under both full and linear finetuning.

\input{tables/rn50_linear_longruns}
\input{tables/rn50_full_longruns}

\section{Experiments on ResNet18 and ResNet34}
\label{appendix:rn18_rn34}

In this section, we further validate our findings for linear finetuning from ResNet50 on two additional smaller architectures, namely ResNet18 and ResNet34. Specifically, we test whether regularization pruning methods generally have better transfer potential than progressive sparsification methods, and whether regularization pruning methods improve over dense models for fine-grained classification tasks. For this purpose, we trained AC/DC and GMP on ImageNet using ResNet18 and ResNet34 models, for 80\% and 90\% sparsity, using the same hyperparameters as for ResNet50. For both ResNet18 and ResNet34, there was a fairly large gap in ImageNet validation accuracy between GMP and AC/DC for both 80\% and 90\% sparsity, in favor of GMP, which almost recovered the baseline accuracy at 80\% sparsity.

We show the results for linear finetuning using AC/DC and GMP in Table~\ref{table:rn18_linear} for ResNet18, respectively Table~\ref{table:rn34_linear} for ResNet34. Interestingly, despite the larger gap in ImageNet validation accuracy between GMP and AC/DC (with GMP being closer to the dense baseline), AC/DC tends to outperform GMP in terms of transfer performance, on most of the downstream tasks. Furthermore, we observe that AC/DC tends to transfer better than the dense baseline, especially for specialized or fine-grained downstream tasks. These observations confirm our findings for linear finetuning on ResNet50.

\begin{table}[h]
\centering
\begin{tabular}{cccccc}
\toprule
{Pruning Strategy} & {Dense} & {GMP 80\%} & {GMP 90\%} & {AC/DC 80\%} & {AC/DC 90\%} \\
{Task} & {} & {} & {} & {} & {} \\
\midrule
Aircraft & 47.7 $ \pm $ 0.1 & 45.5 $ \pm $ 0.1 & 45.6 $ \pm $ 0.1 & 48.0 $ \pm $ 0.1 & \textbf{48.1 $ \pm $ 0.1} \\
Birds & 49.4 $ \pm $ 0.1 & 49.3 $ \pm $ 0.1 & 48.1 $ \pm $ 0.0 & \textbf{50.2 $ \pm $ 0.0} & 48.7 $ \pm $ 0.1 \\
CIFAR-10 & 87.2 $ \pm $ 0.0 & \textbf{87.4 $ \pm $ 0.0} & 87.2 $ \pm $ 0.0 & \textbf{87.4 $ \pm $ 0.0} & 87.2 $ \pm $ 0.1 \\
CIFAR-100 & 68.9 $ \pm $ 0.0 & 68.1 $ \pm $ 0.0 & 69.1 $ \pm $ 0.0 & \textbf{69.6 $ \pm $ 0.1} & 68.9 $ \pm $ 0.0 \\
Caltech-101 & 89.4 $ \pm $ 0.3 & \textbf{89.8 $ \pm $ 0.3} & 88.6 $ \pm $ 0.2 & 89.0 $ \pm $ 0.2 & 88.2 $ \pm $ 0.4 \\
Caltech-256 & \textbf{79.4 $ \pm $ 0.1} & 78.3 $ \pm $ 0.1 & 77.3 $ \pm $ 0.1 & 78.8 $ \pm $ 0.1 & 77.3 $ \pm $ 0.1 \\
Cars & 45.6 $ \pm $ 0.1 & 45.0 $ \pm $ 0.1 & 44.4 $ \pm $ 0.1 & 46.2 $ \pm $ 0.1 & \textbf{46.7 $ \pm $ 0.1} \\
DTD & 68.1 $ \pm $ 0.1 & 68.2 $ \pm $ 0.3 & 66.9 $ \pm $ 0.2 & \textbf{68.6 $ \pm $ 0.2} & 68.4 $ \pm $ 0.2 \\
Flowers & 89.0 $ \pm $ 0.1 & 89.3 $ \pm $ 0.1 & 89.3 $ \pm $ 0.1 & 89.9 $ \pm $ 0.1 & \textbf{90.2 $ \pm $ 0.1} \\
Food-101 & 64.9 $ \pm $ 0.0 & 65.0 $ \pm $ 0.0 & 64.6 $ \pm $ 0.0 & \textbf{65.6 $ \pm $ 0.0} & 65.3 $ \pm $ 0.0 \\
Pets & \textbf{90.1 $ \pm $ 0.1} & 89.8 $ \pm $ 0.1 & 89.4 $ \pm $ 0.2 & 89.7 $ \pm $ 0.1 & 89.4 $ \pm $ 0.1 \\
SUN397 & \textbf{54.8 $ \pm $ 0.1} & 53.8 $ \pm $ 0.1 & 52.9 $ \pm $ 0.1 & \textbf{54.8 $ \pm $ 0.1} & 53.5 $ \pm $ 0.1 \\

\bottomrule
\end{tabular}
\caption{Transfer accuracy for different pruning methods for \emph{linear finetuning} on ResNet18}
\label{table:rn18_linear}
\end{table}

\begin{table}[h]
\centering
\begin{tabular}{cccccc}
\toprule
{Pruning Strategy} & {Dense} & {GMP 80\%} &{GMP 90\%} & {AC/DC 80\%} & {AC/DC 90\%} \\
{Task} & {} & {} & {} & {} \\
\midrule
Aircraft & 45.8 $ \pm $ 0.2 & 43.5 $ \pm $ 0.2 & 44.9 $ \pm $ 0.1 & 48.7 $ \pm $ 0.1 & \textbf{50.7 $ \pm $ 0.2} \\
Birds & 52.9 $ \pm $ 0.0 & 53.0 $ \pm $ 0.1 & 53.0 $ \pm $ 0.1 & \textbf{54.5 $ \pm $ 0.1} & 54.2 $ \pm $ 0.1 \\
CIFAR-10 & 89.5 $ \pm $ 0.0 & 89.1 $ \pm $ 0.0 & 88.5 $ \pm $ 0.0 & \textbf{89.6 $ \pm $ 0.0} & 89.0 $ \pm $ 0.0 \\
CIFAR-100 & 71.0 $ \pm $ 0.0 & 70.4 $ \pm $ 0.1 & 70.2 $ \pm $ 0.1 & \textbf{72.0 $ \pm $ 0.0} & \textbf{72.0 $ \pm $ 0.0} \\
Caltech-101 & \textbf{92.5 $ \pm $ 0.2} & 91.8 $ \pm $ 0.3 & 90.9 $ \pm $ 0.2 & 92.0 $ \pm $ 0.3 & 91.8 $ \pm $ 0.4 \\
Caltech-256 & 82.2 $ \pm $ 0.1 & 81.8 $ \pm $ 0.0 & 81.4 $ \pm $ 0.1 & \textbf{82.3 $ \pm $ 0.1} & 81.2 $ \pm $ 0.1 \\
Cars & 47.3 $ \pm $ 0.1 & 46.0 $ \pm $ 0.1 & 45.6 $ \pm $ 0.1 & 48.5 $ \pm $ 0.1 & \textbf{49.0 $ \pm $ 0.1} \\
DTD & 69.5 $ \pm $ 0.1 & 68.6 $ \pm $ 0.5 & 68.6 $ \pm $ 0.2 & \textbf{70.4 $ \pm $ 0.3} & 69.6 $ \pm $ 0.2 \\
Flowers & 88.1 $ \pm $ 0.1 & 88.5 $ \pm $ 0.1 & 89.0 $ \pm $ 0.1 & 90.0 $ \pm $ 0.1 & \textbf{91.1 $ \pm $ 0.1} \\
Food-101 & 66.8 $ \pm $ 0.0 & 66.7 $ \pm $ 0.0 & 67.4 $ \pm $ 0.0 & 68.2 $ \pm $ 0.0 & \textbf{68.8 $ \pm $ 0.0} \\
Pets & 92.0 $ \pm $ 0.1 & \textbf{92.5 $ \pm $ 0.1} & 91.4 $ \pm $ 0.1 & 91.7 $ \pm $ 0.1 & 91.1 $ \pm $ 0.2 \\
SUN397 & 55.9 $ \pm $ 0.1 & 55.4 $ \pm $ 0.1 & 55.0 $ \pm $ 0.1 & \textbf{56.8 $ \pm $ 0.1} & 55.6 $ \pm $ 0.1 \\

\bottomrule
\end{tabular}
\caption{Transfer accuracy for different pruning methods for \emph{linear finetuning} on ResNet34}
\label{table:rn34_linear}
\end{table}


\section{Experiments on MobileNetV1}
\label{appendix:mobilenet}

The MobileNet \cite{howard2017mobilenets} architecture is a natural choice for devices with limited computational resources. We measure the results of sparse transfer with full and linear finetuning on the same downstream tasks starting from dense ImageNet models pruned using regularization-based and progressive sparsification methods. Specifically, we use AC/DC, STR for regularization methods and M-FAC \cite{frantar2021efficient} for the progressive sparsification category. 

M-FAC is a framework for efficiently computing high-dimensional inverse-Hessian vector products, which can be applied to different scenarios that use second-order information. In particular, one such instance is pruning, where M-FAC aims to solve the same optimization problem as WoodFisher, and thus from this point of view these methods are very similar. In particular, it has been shown \cite{frantar2021efficient} that M-FAC outperforms WoodFisher on ImageNet models, in terms of accuracy at a given sparsity level. Specifically, for MobileNet, M-FAC surpasses all existing methods at 90\% sparsity, reaching 67.2\% validation accuracy. For this reason, we included M-FAC, in favor of WoodFisher, to our list of progressive sparsification methods for MobileNetV1.  

Due to the smaller size of the MobileNetV1 architecture, we additionally test the effect that lower sparsity levels have on the transfer performance, by training on ImageNet AC/DC models at 30\%, 40\% and 50\% sparsity; these models fully recover the dense baseline accuracy on ImageNet.

\begin{figure}
    \centering
    \includegraphics[width=0.9\textwidth]{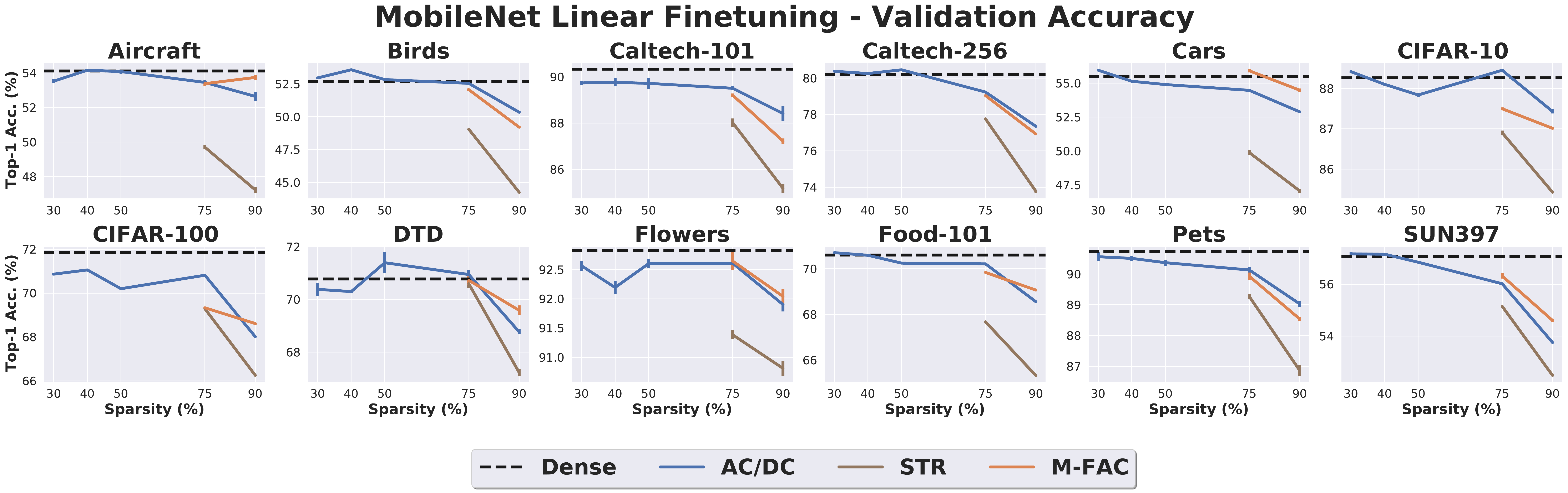}
    \caption{(MobileNetV1) Per-dataset downstream validation accuracy for transfer learining with linear finetuning.}
    \label{fig:individual_mobnet_linear}
\end{figure}

\begin{figure}
    \centering
    \includegraphics[width=0.9\textwidth]{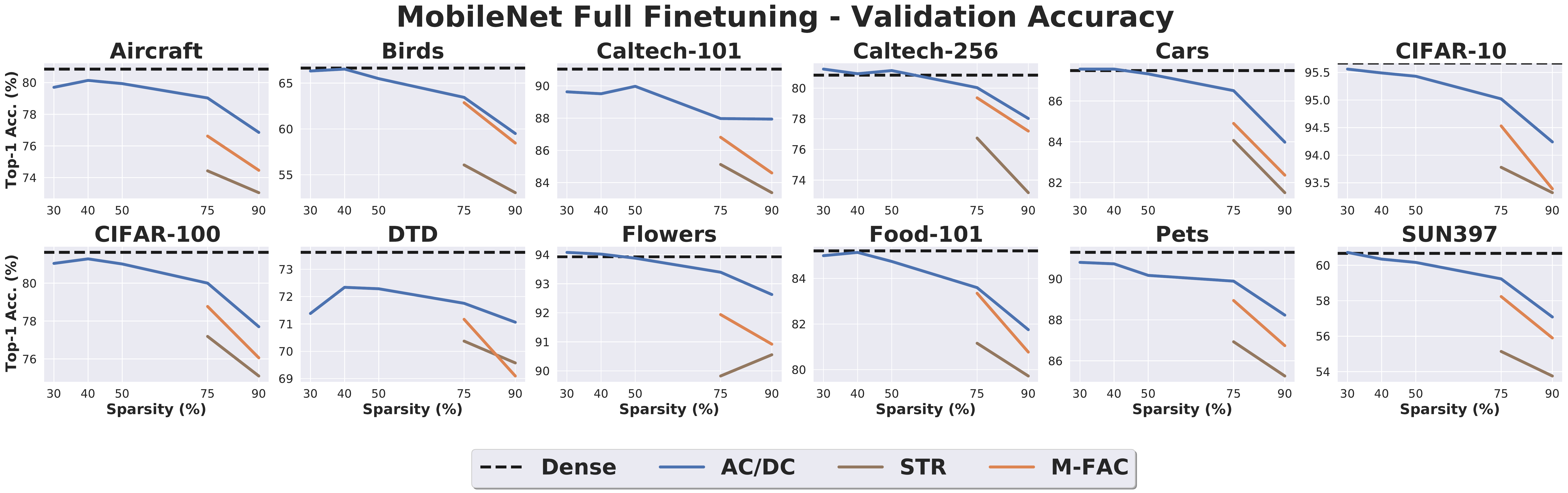}
    \caption{(MobileNetV1) Per-dataset downstream validation accuracy for transfer learining with full finetuning.}
    \label{fig:individual_mobnet_full}
\end{figure}

The results on MobileNet are presented in Figure~\ref{fig:individual_mobnet_linear} and Table~\ref{table:mobilenet_linear} for linear finetuning  and Figure~\ref{fig:individual_mobnet_full} and Table~\ref{table:mobilenet_full} for full finetuning. The results for linear finetuning are obtained after running from five different random seeds, and the mean and standard deviation are reported. However, the experiments for full finetuning were each run once. For both linear and full finetuning, we observe that generally the performance decays faster with increased sparsity, compared to ResNet50; this is expected, given the lower parameter count for MobileNet and the larger gap in ImageNet validation accuracy between dense and sparse models. 

For linear finetuning, we observe AC/DC outperforms STR at both 75\% and 90\% sparsity. Furthermore, AC/DC tends to be close to M-FAC at 75\% sparsity, while at 90\% sparsity M-FAC performs better on almost half of the tasks. Differently from ResNet50, for MobileNet neither regularization based nor progressive sparsification models outperform the dense baseline, at higher sparsity (75\% and 90\%). We observe at lower sparsity (30\% and 50\%) a few instances where sparse models slightly outperform the dense baseline (Birds, Cars, DTD), but generally the differences are not significant.

In the case of full finetuning, we observe that the performance of sparse models decays more quickly than for ResNet50, and even at lower sparsity (30-50\%) there is a gap in transfer performance compared to the dense baseline. Furthermore, AC/DC outperforms STR and M-FAC at both 75\% and 90\% sparsity on all downstream tasks. Overall, the results for MobileNet indicate that the transfer performance is significantly affected by the sparsity of the backbone model, for both linear and full finetuning. Moreover, the experiments on MobileNet seem to suggest that although some of the conclusions derived from the ResNet experiments are confirmed (e.g. sparse models usually have similar or slightly better performance to the dense baseline for linear finetuning), the guidelines for the preferred sparsity method in a given scenario might be specific to the choice of the backbone architecture.  

\begin{table}[h]
\centering
\scalebox{0.8}{
\begin{tabular}{ccccccccccc}
\toprule
{Pruning Strategy} & {Dense} & \multicolumn{1}{c}{\begin{tabular}[c]{@{}c@{}} AC/DC \\ 30\% \end{tabular}} & {\begin{tabular}[c]{@{}c@{}} AC/DC \\ 40\% \end{tabular}} & {\begin{tabular}[c]{@{}c@{}} AC/DC \\ 50\% \end{tabular}} & {\begin{tabular}[c]{@{}c@{}} AC/DC \\ 75\% \end{tabular}} & {\begin{tabular}[c]{@{}c@{}} AC/DC \\ 90\% \end{tabular}} & {\begin{tabular}[c]{@{}c@{}} M-FAC \\ 75\% \end{tabular}} & {\begin{tabular}[c]{@{}c@{}} M-FAC \\ 89\% \end{tabular}} & {\begin{tabular}[c]{@{}c@{}} STR \\ 75\% \end{tabular}} & {\begin{tabular}[c]{@{}c@{}} STR \\ 90\% \end{tabular}} \\
{Task} & {} & {} & {} & {} & {} & {} & {} & {} & {} & {} \\
\midrule
Aircraft & 54.1 $ \pm $ 0.2 & 53.5 $ \pm $ 0.1 & \textbf{54.2 $ \pm $ 0.1} & 54.1 $ \pm $ 0.1 & 53.5 $ \pm $ 0.2 & 52.6 $ \pm $ 0.3 & 53.4 $ \pm $ 0.1 & 53.7 $ \pm $ 0.1 & 49.7 $ \pm $ 0.1 & 47.2 $ \pm $ 0.2 \\
Birds & 52.7 $ \pm $ 0.1 & 53.0 $ \pm $ 0.1 & \textbf{53.6 $ \pm $ 0.1} & 52.8 $ \pm $ 0.0 & 52.6 $ \pm $ 0.1 & 50.3 $ \pm $ 0.1 & 52.1 $ \pm $ 0.1 & 49.2 $ \pm $ 0.1 & 49.0 $ \pm $ 0.0 & 44.2 $ \pm $ 0.0 \\
CIFAR-10 & 88.3 $ \pm $ 0.1 & 88.4 $ \pm $ 0.0 & 88.1 $ \pm $ 0.0 & 87.8 $ \pm $ 0.0 & \textbf{88.5 $ \pm $ 0.0} & 87.4 $ \pm $ 0.1 & 87.5 $ \pm $ 0.0 & 87.0 $ \pm $ 0.0 & 86.9 $ \pm $ 0.1 & 85.4 $ \pm $ 0.0 \\
CIFAR-100 & \textbf{71.9 $ \pm $ 0.0} & 70.9 $ \pm $ 0.1 & 71.1 $ \pm $ 0.0 & 70.2 $ \pm $ 0.0 & 70.8 $ \pm $ 0.0 & 68.0 $ \pm $ 0.0 & 69.3 $ \pm $ 0.0 & 68.6 $ \pm $ 0.0 & 69.3 $ \pm $ 0.0 & 66.3 $ \pm $ 0.0 \\
Caltech-101 & \textbf{90.3 $ \pm $ 0.1} & 89.7 $ \pm $ 0.1 & 89.8 $ \pm $ 0.2 & 89.7 $ \pm $ 0.2 & 89.5 $ \pm $ 0.1 & 88.4 $ \pm $ 0.3 & 89.2 $ \pm $ 0.1 & 87.2 $ \pm $ 0.1 & 88.0 $ \pm $ 0.2 & 85.2 $ \pm $ 0.2 \\
Caltech-256 & 80.2 $ \pm $ 0.1 & 80.4 $ \pm $ 0.0 & 80.2 $ \pm $ 0.1 & \textbf{80.5 $ \pm $ 0.0} & 79.2 $ \pm $ 0.0 & 77.3 $ \pm $ 0.1 & 79.0 $ \pm $ 0.0 & 76.9 $ \pm $ 0.0 & 77.8 $ \pm $ 0.1 & 73.8 $ \pm $ 0.1 \\
Cars & 55.5 $ \pm $ 0.0 & \textbf{55.9 $ \pm $ 0.1} & 55.1 $ \pm $ 0.1 & 54.9 $ \pm $ 0.1 & 54.5 $ \pm $ 0.1 & 52.9 $ \pm $ 0.0 & \textbf{55.9 $ \pm $ 0.1} & 54.5 $ \pm $ 0.1 & 49.9 $ \pm $ 0.2 & 47.1 $ \pm $ 0.1 \\
DTD & 70.8 $ \pm $ 0.2 & 70.4 $ \pm $ 0.2 & 70.3 $ \pm $ 0.0 & \textbf{71.4 $ \pm $ 0.4} & 70.9 $ \pm $ 0.2 & 68.8 $ \pm $ 0.1 & 70.7 $ \pm $ 0.2 & 69.6 $ \pm $ 0.2 & 70.6 $ \pm $ 0.2 & 67.2 $ \pm $ 0.1 \\
Flowers & \textbf{92.8 $ \pm $ 0.1} & 92.6 $ \pm $ 0.1 & 92.2 $ \pm $ 0.1 & 92.6 $ \pm $ 0.1 & 92.6 $ \pm $ 0.1 & 91.9 $ \pm $ 0.1 & 92.6 $ \pm $ 0.1 & 92.0 $ \pm $ 0.1 & 91.4 $ \pm $ 0.1 & 90.8 $ \pm $ 0.1 \\
Food-101 & 70.6 $ \pm $ 0.0 & \textbf{70.7 $ \pm $ 0.0} & 70.6 $ \pm $ 0.0 & 70.3 $ \pm $ 0.0 & 70.2 $ \pm $ 0.0 & 68.6 $ \pm $ 0.0 & 69.8 $ \pm $ 0.0 & 69.1 $ \pm $ 0.0 & 67.7 $ \pm $ 0.0 & 65.3 $ \pm $ 0.0 \\
Pets & \textbf{90.7 $ \pm $ 0.1} & 90.6 $ \pm $ 0.1 & 90.5 $ \pm $ 0.1 & 90.4 $ \pm $ 0.1 & 90.1 $ \pm $ 0.1 & 89.0 $ \pm $ 0.1 & 89.9 $ \pm $ 0.1 & 88.5 $ \pm $ 0.1 & 89.3 $ \pm $ 0.1 & 86.9 $ \pm $ 0.2 \\
SUN397 & 57.1 $ \pm $ 0.0 & \textbf{57.2 $ \pm $ 0.1} & 57.2 $ \pm $ 0.0 & 56.8 $ \pm $ 0.0 & 56.0 $ \pm $ 0.0 & 53.8 $ \pm $ 0.0 & 56.3 $ \pm $ 0.1 & 54.6 $ \pm $ 0.0 & 55.1 $ \pm $ 0.0 & 52.5 $ \pm $ 0.0 \\
\bottomrule
\end{tabular}
}
\caption{Transfer accuracy for \emph{linear finetuning} using sparse MobileNet models}
\label{table:mobilenet_linear}
\end{table}

\begin{table}[h]
\centering
\scalebox{0.9}{
\begin{tabular}{ccccccccccc}
\toprule
{Pruning Strategy} & {Dense} & \multicolumn{1}{c}{\begin{tabular}[c]{@{}c@{}} AC/DC \\ 30\% \end{tabular}} & {\begin{tabular}[c]{@{}c@{}} AC/DC \\ 40\% \end{tabular}} & {\begin{tabular}[c]{@{}c@{}} AC/DC \\ 50\% \end{tabular}} & {\begin{tabular}[c]{@{}c@{}} AC/DC \\ 75\% \end{tabular}} & {\begin{tabular}[c]{@{}c@{}} AC/DC \\ 90\% \end{tabular}} & {\begin{tabular}[c]{@{}c@{}} M-FAC \\ 75\% \end{tabular}} & {\begin{tabular}[c]{@{}c@{}} M-FAC \\ 89\% \end{tabular}} & {\begin{tabular}[c]{@{}c@{}} STR \\ 75\% \end{tabular}} & {\begin{tabular}[c]{@{}c@{}} STR \\ 90\% \end{tabular}} \\
\midrule
Aircraft & \textbf{80.9} & 79.7 & 80.1 & 79.9 & 79.0 & 76.9 & 76.6 & 74.5 & 74.4 & 73.0 \\
Birds & \textbf{66.6} & 66.3 & 66.5 & 65.5 & 63.4 & 59.5 & 62.9 & 58.5 & 56.1 & 53.1 \\
CIFAR-10 & \textbf{95.7} & 95.6 & 95.5 & 95.4 & 95.0 & 94.2 & 94.5 & 93.4 & 93.8 & 93.3 \\
CIFAR-100 & \textbf{81.6} & 81.0 & 81.3 & 81.0 & 80.0 & 77.7 & 78.8 & 76.1 & 77.2 & 75.1 \\
Caltech-101 & \textbf{91.0} & 89.6 & 89.5 & 90.0 & 88.0 & 87.9 & 86.8 & 84.6 & 85.1 & 83.4 \\
Caltech-256 & 80.9 & \textbf{81.2} & 80.9 & 81.1 & 80.0 & 78.0 & 79.4 & 77.2 & 76.7 & 73.2 \\
Cars & 87.5 & \textbf{87.6} & \textbf{87.6} & 87.3 & 86.5 & 84.0 & 84.9 & 82.4 & 84.1 & 81.5 \\
DTD & \textbf{73.6} & 71.4 & 72.3 & 72.3 & 71.8 & 71.1 & 71.2 & 69.1 & 70.4 & 69.6 \\
Flowers & 93.9 & \textbf{94.1} & 94.0 & 93.9 & 93.4 & 92.6 & 91.9 & 90.9 & 89.8 & 90.6 \\
Food-101 & \textbf{85.2} & 85.0 & 85.1 & 84.7 & 83.6 & 81.8 & 83.4 & 80.8 & 81.2 & 79.7 \\
Pets & \textbf{91.3} & 90.8 & 90.7 & 90.2 & 89.9 & 88.2 & 88.9 & 86.7 & 86.9 & 85.3 \\
SUN397 & \textbf{60.7} & \textbf{60.7} & 60.3 & 60.2 & 59.2 & 57.1 & 58.2 & 55.9 & 55.1 & 53.8 \\
\bottomrule
\end{tabular}
}
\caption{Transfer accuracy for \emph{full finetuning} using sparse MobileNet models}
\label{table:mobilenet_full}
\end{table}

Finally, we consider the accuracy tradeoff of using a smaller network such as MobileNet (4.2M trainable weights) versus a larger model, ResNet50 (25.5M trainable weights), but pruned to 90\% sparsity. We present linear and full finetuning accuracy results for these two scenarios for an easier comparison in Tables~\ref{table:comparison_rn50_mobnet_linear} and~\ref{table:comparison_rn50_mobnet_full}. We use the overall best pruning strategy for each type of transfer on ResNet50: AC/DC for linear finetuning and WoodFisher for full finetuning. Note that these same results are also presented in Tables \ref{table:rn50_full_all} and \ref{table:rn50_linear_all}, \ref{table:mobilenet_full}, and \ref{table:mobilenet_linear}.

We observe that generally, pruning ResNet50 to 80 or even 90\% sparsity results in higher accuracy than MobileNet, for both linear and full finetuning. However, in almost all cases, the gap is below 5\%. This finding confirms conventional wisdom that training and pruning large networks generally results in higher accuracy than training dense small networks from scratch.

\begin{table}[h]
\begin{minipage}[t]{.47\textwidth}
\centering
\scalebox{0.9}{
\begin{tabular}{cccc}
\toprule
{Model} & 
 \multicolumn{1}{c}{
 \begin{tabular}[c]{@{}c@{}} MobileNet \\ Dense \end{tabular}} & {\begin{tabular}[c]{@{}c@{}} ResNet50 \\ AC/DC 80\% \end{tabular}} &  {\begin{tabular}[c]{@{}c@{}} ResNet50 \\ AC/DC 90\% \end{tabular}} \\
\midrule
Aircraft & 54.1 $ \pm $ 0.2 & 55.1 $\pm$ 0.1 & 55.5 $\pm$ 0.1  \\
Birds & 52.7 $ \pm $ 0.1 & 58.4 $\pm$ 0.0 & 58.7 $\pm$ 0.0\\
CIFAR-10 & 88.3 $ \pm $ 0.1  & 90.9 $\pm$ 0.0 & 91.0 $\pm$ 0.0 \\
CIFAR-100 & 71.9 $ \pm $ 0.0 & 74.7 $\pm$ 0.1 & 74.3 $\pm$ 0.0  \\
Caltech-101 & 90.3 $ \pm $ 0.1 & 92.4 $\pm$ 0.2 & 92.5 $\pm$ 0.1  \\
Caltech-256 & 80.2 $ \pm $ 0.1 & 84.6 $\pm$ 0.1 & 84.5 $\pm$ 0.0 \\
Cars & 55.5 $\pm$ 0.0 & 56.6 $\pm$ 0.0 & 56.0 $\pm$ 0.1 \\
DTD & 70.8 $ \pm $ 0.2 & 74.4 $\pm$ 0.1 & $73.7 \pm$ 0.2 \\
Flowers & 92.8 $ \pm $ 0.1 & 92.7 $\pm$ 0.1 & 92.4 $\pm$ 0.0 \\
Food-101 & 70.6 $ \pm $ 0.0  & 73.8 $\pm$ 0.0 & 73.8 $\pm$ 0.0 \\
Pets & 90.7 $ \pm $ 0.1 & 92.3 $\pm$ 0.1 & 91.9 $\pm$ 0.1 \\
SUN397 & 57.1 $ \pm $ 0.0 & 60.4 $\pm$ 0.0 & 59.8 $\pm$ 0.1 \\
\bottomrule
\end{tabular}
}
\caption{Comparison of MobileNet dense versus\\ ResNet50 sparse models when transferring with \emph{linear\\ finetuning}}
\label{table:comparison_rn50_mobnet_linear}
\end{minipage}%
\hspace{0.7cm}
\begin{minipage}[t]{.47\textwidth}
\centering
\scalebox{0.9}{
\begin{tabular}{cccc}
\toprule
{Model} & 
 \multicolumn{1}{c}{
 \begin{tabular}[c]{@{}c@{}} MobileNet \\ Dense \end{tabular}} & {\begin{tabular}[c]{@{}c@{}} ResNet50 \\ WoodFisher 80\% \end{tabular}} &  {\begin{tabular}[c]{@{}c@{}} ResNet50 \\ WoodFisher 90\% \end{tabular}} \\
\midrule
Aircraft & 80.9 & 84.8 $\pm$ 0.2 & 84.5 $\pm$ 0.4  \\
Birds & 66.6 & 72.4 $\pm$ 0.4 & 71.6 $\pm$ 0.2  \\
CIFAR-10 & 95.7  & 97.2 $\pm$ 0.1  & 97.0 $\pm$ 0.1  \\
CIFAR-100 & 81.6 & 85.1 $\pm$ 0.1  & 84.4 $\pm$ 0.2 \\
Caltech-101 & 91.0 & 93.7 $\pm$ 0.1  & 93.9 $\pm$ 0.3 \\
Caltech-256 & 80.9 & 85.1 $\pm$ 0.1  & 84.0$\pm$ 0.1  \\
Cars & 87.5 & 90.5 $\pm$ 0.2  & 90.0 $\pm$ 0.2  \\
DTD & 73.6 & 75.4 $\pm$ 0.3  & 75.5 $\pm$ 0.4  \\
Flowers & 93.9 & 95.5 $\pm$ 0.2  & 95.5 $\pm$ 0.3  \\
Food-101 & 85.2  & 87.4 $\pm$ 0.1  & 87.0 $\pm$ 0.1 \\
Pets & 91.3 & 93.3 $\pm$ 0.3 & 92.7 $\pm$ 0.3 \\
SUN397 & 60.7 & 62.8 $\pm$ 0.1 & 62.3 $\pm$ 0.1 \\
\bottomrule
\end{tabular}
}
\caption{Comparison of MobileNet dense versus ResNet50 sparse models when transferring with \emph{full finetuning}}
\label{table:comparison_rn50_mobnet_full}
\end{minipage}%
\end{table}

\clearpage

\section{Impact of fully connected layer bias on full finetuning transfer accuracy}
In our experiments, we used the original architectures used to train the upstream ImageNet models when performing transfer with full-finetuning, only resizing the final layer to match the number of output classes in the downstream task. This choice was necessitated partially by ensuring that the weights were applied correctly. For example, the RigL models were trained using TensorFlow, which uses slightly different Convolution and MaxPooling padding conventions than PyTorch. Likewise, STR models were trained using a slightly nonstandard PyTorch implementation of ResNet50, which did not use a bias term in the final Fully-Connected (FC) layer. We investigate the possibility that the latter difference could have an effect on downstream transfer accuracy. To do so, we transferred a set of 80\% sparse ResNet50 STR models to all downstream tasks, using a bias term in the FC layer. The results are shown in Table~\ref{table:fcbias-summary}. Additionally, we perform a similar comparison on MobileNetV1, for STR models at 75\% sparsity. As in the case of ResNet50, the version of MobileNet used by the STR models does not use bias in the final classification layer. The results illustrating the bias effect on full finetuning for MobileNet are presented in Table~
\ref{table:fcbias-summary-mobnet}. We observe that the presence of a bias term in the final layer can, in some cases, have a small positive effect on the resulting model, and so we caution that these effects be considered when choosing a transfer architecture. 

\begin{table}[h]
\begin{minipage}{0.47\textwidth}
\centering

\scalebox{0.9}{%
\begin{tabular}{ccc}
\toprule
{Dataset} & {With FC Bias} & {Without FC Bias} \\
\midrule
Aircraft & 79.8 $ \pm $ 0.6 & 79.8 $ \pm $ 0.3 \\
Birds & 67.9 $ \pm $ 0.2 & 68.1 $ \pm $ 0.1 \\
CIFAR-10 & 96.5 $ \pm $ 0 & 96.5 $ \pm $ 0.1 \\
CIFAR-100 & 83.7 $ \pm $ 0.2 & 83.6 $ \pm $ 0.2 \\
Caltech-101 & 91.2 $ \pm $ 0.2 & 90.7 $ \pm $ 0.6 \\
Caltech-256 & 84.4 $ \pm $ 0.1 & 84.0 $ \pm $ 0.1 \\
Cars & 87.7 $ \pm $ 0.1 & 87.8 $ \pm $ 0.1 \\
DTD & 74.4 $ \pm $ 0.2 & 73.7 $ \pm $ 0.6 \\
Flowers & 94 $ \pm $ 0.1 & 93.7 $ \pm $ 0.2 \\
Food-101 & 86 $ \pm $ 0.1 & 85.9 $ \pm $ 0.1 \\
Pets & 92.1 $ \pm $ 0.1 & 92.1 $ \pm $ 0.1 \\
SUN397 & 63.2 $ \pm $ 0.1 & 62.6 $ \pm $ 0.1 \\

\bottomrule
\end{tabular}
}
\caption{Top-1 validation accuracy on ResNet50 trained using STR on ImageNet, with using bias in the FC layer versus without. The original model architecture does not use bias in the FC layer.}
\label{table:fcbias-summary}
\end{minipage}
\hspace{0.3cm}
\begin{minipage}{0.47\textwidth}
\centering
\scalebox{0.9}{
\begin{tabular}{ccc}
\toprule
{Dataset}  & {With FC Bias}  &  {Without FC Bias}\\
\midrule
Aircraft  & 74.2  &  74.4\\
Birds & 56.4 &  56.1  \\
CIFAR-10 & 93.9 &  93.8  \\
CIFAR-100 & 77.5  &  77.2  \\
Caltech-101 & 86.3 &  85.1 \\
Caltech-256 & 76.9 &  76.7  \\
Cars & 83.8 &  84.1  \\
DTD & 71.8 &  70.4  \\
Flowers & 90.4 &  89.8  \\
Food-101 & 80.9 & 81.2  \\
Pets & 87.9 &  86.9  \\
SUN397 & 56.1 & 55.1  \\
\bottomrule
\end{tabular}}
\caption{Top-1 validation accuracy on MobileNetV1 trained using STR on ImageNet, with bias in the FC layer versus without. The original model architecture does not use bias in the FC layer.}
\label{table:fcbias-summary-mobnet}
\end{minipage}
\end{table}

\section{Impact of label smoothing on transfer accuracy}
\label{section:label_smoothing}
 We take advantage of the fact that we have STR checkpoints trained with and without label smoothing (LS) to investigate the effect of LS on dense and sparse transfer accuracy in the context of linear transfer.
 As Table \ref{table:str_ls} shows, label smoothing tends to have a negative effect on transfer accuracy (confirming the results in \cite{kornblith2019better}). However, our experiments suggest that this effect is more pronounced on the Aircraft and Cars datasets in the case of sparse STR models, and generally for most specialized datasets for the dense models. Furthermore, we observe that the performance gap tends to narrow with increased sparsity. We also note that even with label smoothing, at 80\% sparsity STR matches or outperforms GMP on all datasets, although the effect largely reverses at 90\% sparsity. 
 
 Overall, these data can be taken as a preliminary confirmation of the importance of controlling for variation in hyperparameters when comparing the transfer performance of various training and pruning methods.

\begin{table}[h]
\centering
\scalebox{0.75}{
\begin{tabular}{c|cc|cc|cc|cc|cc}
\toprule
{} & {Dense} & {Dense LS} & {STR 80\%} & {STR LS 80\%} & {STR 90\%} & {STR LS 90\%} & {STR 95\%} & {STR LS 95\%} & {STR 98\%} & {STR LS 98\%} \\
{Dataset} & {} & {} & {} & {} & {} & {} & {} & {} & {} & {} \\
\midrule
Aircraft & 49.2 $ \pm $ 0.1 & 38.2 $ \pm $ 0.1 & 53.7 $ \pm $ 0.0 & 47.0 $ \pm $ 0.0 & 52.9 $ \pm $ 0.1 & 46.4 $ \pm $ 0.1 & 50.3 $ \pm $ 0.1 & 46.6 $ \pm $ 0.1 & 48.0 $ \pm $ 0.1 & 45.2 $ \pm $ 0.1 \\
Birds & 57.7 $ \pm $ 0.1 & 52.4 $ \pm $ 0.0 & 56.2 $ \pm $ 0.1 & 56.4 $ \pm $ 0.0 & 55.2 $ \pm $ 0.1 & 56.0 $ \pm $ 0.0 & 52.1 $ \pm $ 0.1 & 51.7 $ \pm $ 0.1 & 43.7 $ \pm $ 0.0 & 45.6 $ \pm $ 0.0 \\
CIFAR-10 & 91.2 $ \pm $ 0.0 & 89.6 $ \pm $ 0.0 & 91.4 $ \pm $ 0.0 & 90.1 $ \pm $ 0.0 & 90.6 $ \pm $ 0.0 & 89.4 $ \pm $ 0.0 & 89.1 $ \pm $ 0.0 & 88.6 $ \pm $ 0.0 & 86.5 $ \pm $ 0.0 & 86.0 $ \pm $ 0.0 \\
CIFAR-100 & 74.6 $ \pm $ 0.1 & 71.6 $ \pm $ 0.0 & 74.7 $ \pm $ 0.0 & 73.3 $ \pm $ 0.0 & 73.7 $ \pm $ 0.1 & 72.2 $ \pm $ 0.1 & 71.7 $ \pm $ 0.0 & 70.1 $ \pm $ 0.0 & 67.4 $ \pm $ 0.0 & 66.3 $ \pm $ 0.0 \\
Caltech-101 & 91.9 $ \pm $ 0.1 & 91.6 $ \pm $ 0.1 & 91.2 $ \pm $ 0.1 & 92.6 $ \pm $ 0.1 & 90.9 $ \pm $ 0.1 & 91.1 $ \pm $ 0.2 & 90.0 $ \pm $ 0.2 & 89.8 $ \pm $ 0.1 & 86.3 $ \pm $ 0.1 & 85.4 $ \pm $ 0.1 \\
Caltech-256 & 84.8 $ \pm $ 0.1 & 84.6 $ \pm $ 0.1 & 83.6 $ \pm $ 0.0 & 84.3 $ \pm $ 0.0 & 82.6 $ \pm $ 0.0 & 82.6 $ \pm $ 0.1 & 80.2 $ \pm $ 0.1 & 79.7 $ \pm $ 0.0 & 73.4 $ \pm $ 0.1 & 73.8 $ \pm $ 0.0 \\
Cars & 53.4 $ \pm $ 0.1 & 44.9 $ \pm $ 0.1 & 57.0 $ \pm $ 0.1 & 50.9 $ \pm $ 0.0 & 54.8 $ \pm $ 0.1 & 49.8 $ \pm $ 0.1 & 50.5 $ \pm $ 0.1 & 46.9 $ \pm $ 0.1 & 44.4 $ \pm $ 0.1 & 42.5 $ \pm $ 0.1 \\
DTD & 73.5 $ \pm $ 0.2 & 72.3 $ \pm $ 0.1 & 74.3 $ \pm $ 0.2 & 73.9 $ \pm $ 0.3 & 73.8 $ \pm $ 0.1 & 73.7 $ \pm $ 0.2 & 72.1 $ \pm $ 0.2 & 71.9 $ \pm $ 0.1 & 68.4 $ \pm $ 0.2 & 68.3 $ \pm $ 0.1 \\
Flowers & 91.6 $ \pm $ 0.1 & 86.7 $ \pm $ 0.1 & 93.0 $ \pm $ 0.0 & 91.2 $ \pm $ 0.0 & 93.0 $ \pm $ 0.1 & 92.1 $ \pm $ 0.1 & 91.9 $ \pm $ 0.1 & 91.0 $ \pm $ 0.1 & 90.8 $ \pm $ 0.1 & 90.4 $ \pm $ 0.1 \\
Food-101 & 73.2 $ \pm $ 0.0 & 69.5 $ \pm $ 0.0 & 73.9 $ \pm $ 0.0 & 72.2 $ \pm $ 0.0 & 72.6 $ \pm $ 0.0 & 71.1 $ \pm $ 0.0 & 70.7 $ \pm $ 0.0 & 68.8 $ \pm $ 0.0 & 65.3 $ \pm $ 0.0 & 64.3 $ \pm $ 0.0 \\
Pets & 92.6 $ \pm $ 0.1 & 92.9 $ \pm $ 0.1 & 91.7 $ \pm $ 0.0 & 92.4 $ \pm $ 0.1 & 91.1 $ \pm $ 0.1 & 91.7 $ \pm $ 0.1 & 89.8 $ \pm $ 0.1 & 90.1 $ \pm $ 0.1 & 85.5 $ \pm $ 0.1 & 86.6 $ \pm $ 0.1 \\
SUN397 & 60.1 $ \pm $ 0.0 & 59.3 $ \pm $ 0.1 & 60.3 $ \pm $ 0.0 & 60.0 $ \pm $ 0.1 & 58.2 $ \pm $ 0.0 & 58.5 $ \pm $ 0.1 & 56.3 $ \pm $ 0.0 & 55.8 $ \pm $ 0.0 & 50.9 $ \pm $ 0.0 & 51.0 $ \pm $ 0.0 \\
\bottomrule

\end{tabular}
}
\caption{Linear Finetuning Validation Accuracy of STR-pruned and dense models with and without label smoothing.}
\label{table:str_ls}
\end{table}

\section{Finetuning with structured sparsity}
\label{appendix:structured_sparsity}

In this section, we examine the transfer properties of models that were sparsified using structured pruning methods, which remove entire convolutional filters. Specifically, we use both ResNet50 and MobileNetV1 models trained on ImageNet and we do full finetuning on all twelve downstream tasks.

\subsection{ResNet50 with structured sparsity}

We consider a ResNet50 model that was pruned with progressive sparsification, using the $L_1$ magnitude of the convolutional filters as a pruning criterion. The resulting model has an ImageNet validation accuracy of 75.7\% and results in 2.2x inference speed-up compared to the dense baseline, when evaluated on a single sample; this makes it comparable to unstructured $90\%$ sparse models that achieve a similar inference speed-up (please see Table~\ref{table:train-speed}).
The results for full finetuning with the structured sparse model, together with the best results for  dense and unstructured 80\% and 90\% models are presented in Table~\ref{table:rn50-structured}. We observe that models with structured sparsity transfer similarly to or worse than unstructured 90\% sparse models.  Note that the unstructured ResNet50 model has higher ImageNet accuracy compared to 90\% sparse models, at a similar inference speed-up. These results align with the observations made in Section~\ref{subsec:discussion_filters}, that having fewer filters in the structured sparse models limits their capability of expressing features.

\begin{table}[h]
\begin{minipage}[t]{.53\textwidth}
\centering
\scalebox{0.9}{%
\begin{tabular}{ccccc}
\toprule
{Dataset} &  {Dense} & {Structured} & {Best 80\%} & {Best 90\%}  \\
\midrule
Aircraft & $83.6\pm 0.4$ & $81.8 \pm 0.5$ &  \textbf{84.8$\pm$ 0.2} & \textbf{84.9 $\pm$ 0.3} \\
Birds & 72.4 $\pm$ 0.3 & $70.7 \pm 0.1$ & \textbf{73.4$\pm$ 0.1} & $72.9 \pm 0.2$ \\
Caltech101 & \textbf{93.5$\pm$ 0.1} & $92.8 \pm 0.1$ &  \textbf{93.7$\pm$ 0.1} & \textbf{93.9 $\pm$ 0.3} \\
Caltech256 & \textbf{86.1 $\pm$ 0.1} & $84.6 \pm 0.1$ &  $85.4 \pm 0.2$ & $84.8 \pm 0.1$ \\
Cars &  $90.3 \pm 0.2$  & $89.4 \pm 0.0$ & \textbf{90.5 $\pm$ 0.2} & $90.0 \pm 0.2$ \\
CIFAR-10 & \textbf{97.4 $\pm$ 0.} & $97.1 \pm 0.1$  &  $97.2 \pm 0.1$ & $97.1 \pm 0.$ \\
CIFAR-100 &  \textbf{85.6 $\pm$ 0.2} & $84.7 \pm 0.2$ &  $85.1 \pm 0.1$ & $84.4 \pm 0.2$ \\
DTD & \textbf{76.2 $\pm$ 0.3} & $75.2 \pm 0.2$   &  \textbf{75.7 $\pm$ 0.5} & 75.5 $\pm$ 0.4 \\
Flowers & $95.0 \pm 0.1$ &  $95.2 \pm 0.0$ &   \textbf{96.1 $\pm$ 0.1} & \textbf{96.1 $\pm$ 0.1} \\
Food-101 & \textbf{87.3 $\pm$ 0.1} & $86.3 \pm 0.1$  &  \textbf{87.4 $\pm$ 0.1} & \textbf{87.3 $\pm$ 0.2} \\
Pets &  \textbf{93.4 $\pm$ 0.1} & $92.5 \pm 0.1$  &   \textbf{93.4 $\pm$ 0.2} & $92.7 \pm 0.3$ \\
SUN397 & \textbf{64.8 $\pm$ 0.} & $63.4 \pm 0.1$  & $64.0 \pm 0.$ & $63.0 \pm 0.$ \\
\bottomrule
\end{tabular}
}
\caption{(ResNet50) Comparison on full finetuning between \\ dense baseline, models with structured sparsity, and best results \\ for unstructured 80\% and 90\% sparsity.}
\label{table:rn50-structured}
\end{minipage}
\hspace{0.2cm}
\begin{minipage}[t]{.47\textwidth}
\centering
\scalebox{0.9}{%
\begin{tabular}{cccc}
\toprule
{Dataset} &  Dense & 50\% Time &  50\% FLOPs \\
\midrule
Aircraft   &     80.9 &   \textbf{82.9} &     \textbf{83.0} \\
Birds      &    \textbf{66.6} &    66.1 &     66.1 \\
Caltech101 &    \textbf{91.0} &    88.6 &    88.9 \\
Caltech256 &    \textbf{80.9} &    78.6 &    78.4 \\
Cars       &    87.5 &	   \textbf{88.4}	&   \textbf{88.3} \\
CIFAR-10    &   \textbf{95.7} &    95.2 &   95.3 \\
CIFAR-100   &   \textbf{81.6}  &    79.9 &     80.2 \\
DTD        &    \textbf{73.6} &    71.1 &    72.2 \\
Flowers    &    93.9 &    \textbf{94.1} &   \textbf{94.1} \\
Food-101   &    \textbf{85.2} &    84.6 &   84.5 \\
Pets       &    \textbf{91.3} &    91.0 &    91.0 \\
SUN397     &    \textbf{60.7} &  59.4 &  59.1 \\

\bottomrule
\end{tabular}}
\caption{(MobileNet) Full finetuning validation accuracy for MobileNet models with structured sparsity, at 50\% inference time or 50\% inference FLOPs.}
\label{table:mobnet-structured}
\end{minipage}
\end{table}

\subsection{MobileNet with structured sparsity}

We additionally perform full finetuning using MobileNet models pruned for structured sparsity. For these experiments, we use the upstream models provided in \cite{he2018amc}; specifically, we use the MobileNet models that achieve 50\% of the inference time or have 50\% of the dense FLOPs. These models achieve $70.2\%$ and $70.5\%$ ImageNet validation accuracy, respectively. The results presented in Table~\ref{table:mobnet-structured} show that in general models with structured sparsity perform similar to or worse than their dense counterparts, with the exception of Aircraft and Cars where these models significantly outperform the dense baseline.

\section{Sparse Transfer Learning for Segmentation}
\label{appendix:segmentation-sparse}

To complement the  experiments for object detection, we executed transfer learning for a YOLACT model~
\cite{yolact} using a ResNet-101 backbone, that has been trained and sparsified on the segmentation version of the COCO dataset. The average sparsity of the model is $\sim$ 87\%, obtained via gradual magnitude pruning (GMP). The model has mAP@0.5 values 49.36 (bounding box), and 46.37 (mask), versus 50.16 (bounding box), 46.57 (mask) for the dense model on COCO. We transfer the pruned trained weights onto the Pascal dataset. The prediction heads get initialized as dense, and kept dense for transfer. 
The results are presented in Tables~\ref{table:segmentation_dense} and~\ref{table:segmentation_sparse}, and show that indeed sparse transfer is competitive against the dense variant in this case as well. 

\begin{table}[t!]
\centering
\scalebox{0.9}{%
\begin{tabular}{c|ccccccccccc}
\toprule
{Type} & all & 0.5 & 0.55 & 0.6 & 0.65 & 0.7 & 0.75 & 0.8 & 0.85 & 0.9 & 0.95 \\
\midrule
box & 32.62 & 54.05 & 51.96 & 48.72 & 44.81 & 40.84 & 34.72 & 26.89 & 17.33 & 6.33 & 0.55 \\
mask & 30.74 & 50.28 & 47.66 & 44.57 & 41.02 & 36.39 & 31.47 & 25.53 & 18.55 & 10.03 & 1.91 \\
\bottomrule
\end{tabular}
}
\caption{Mean average precision for dense transfer on Pascal, at various thresholds.}
\label{table:segmentation_dense}
\end{table}

\begin{table}[t!]
\centering
\scalebox{0.9}{%
\begin{tabular}{c|ccccccccccc}
\toprule
{Type} & all & 0.5 & 0.55 & 0.6 & 0.65 & 0.7 & 0.75 & 0.8 & 0.85 & 0.9 & 0.95 \\
\midrule
box & 33.55 & 54.15 & 51.79 & 49.2 & 45.57 & 41.51 & 35.95 & 29.2 & 19.66 & 7.78 & 0.74 \\
mask & 31.5 & 50.66 & 47.89 & 45.04 & 41.67 & 37.32 & 32.39 & 26.35 & 19.98 & 11.22 & 2.5 \\
\bottomrule
\end{tabular}
}
\caption{Mean average precision for \emph{sparse} transfer on Pascal, at various thresholds. Notice the similar or slightly improved accuracy.}
\label{table:segmentation_sparse}
\end{table}

\section{Distillation from Sparse Teachers}
\label{appendix:kd-sparse}
Our linear finetuning experiments suggest that sparse models may provide superior representations relative to dense ones. 
To further test this hypothesis, we employ sparse models as teachers in a standard knowledge distillation (KD) setting, 
i.e. training a ResNet34 student model with distillation from a ResNet50 teacher, which may be dense or sparse. 
The accuracy of the resulting models is provided in Table~\ref{table:kd-summary}. 

Results suggest that differences in accuracy between the sparse and dense teachers do not affect distillation. Sparse teachers will also reduce distillation overhead due to faster inference.

\begin{table}[h]
\centering

\scalebox{0.9}{%
\begin{tabular}{ccccccc}
\toprule
{Baseline} & {Dense KD} & \multicolumn{1}{c}{\begin{tabular}[c]{@{}c@{}} AC/DC \\ 80\% \end{tabular}} & \multicolumn{1}{c}{\begin{tabular}[c]{@{}c@{}} WoodFisher \\ 80\% \end{tabular}} & \multicolumn{1}{c}{\begin{tabular}[c]{@{}c@{}} AC/DC \\ 90\% \end{tabular}}  & \multicolumn{1}{c}{\begin{tabular}[c]{@{}c@{}} WoodFisher \\ 90\% \end{tabular}} \\
\midrule
73.83\% & 74.42\% & 74.64\% & 74.63\% & 74.19\% & 74.44\% \\

\bottomrule
\end{tabular}
}
\caption{Top-1 validation accuracy on ResNet34 trained on ImageNet, when distilling from dense or sparse teachers.}
\label{table:kd-summary}
\end{table}

\end{document}

%% file: math_commands.tex
\usepackage{amsmath,amsfonts,bm}

\def\eqref#1{equation~\ref{#1}}

\def\1{\bm{1}}

\DeclareMathAlphabet{\mathsfit}{\encodingdefault}{\sfdefault}{m}{sl}
\SetMathAlphabet{\mathsfit}{bold}{\encodingdefault}{\sfdefault}{bx}{n}



%% file: tables/rn50_linear_longruns.tex
\begin{table}[htb!]
\centering
\scalebox{0.9}{
\begin{tabular}{ccc|ccc|cc}
\toprule
{} & {Dense} & {Dense 2x} & {AC/DC} & {AC/DC 3x} & {AC/DC 5x} & {RigL ERK 1x} & {RigL ERK 5x} \\
\midrule
{80\% Sparsity} & {} & {} & {} & {} & {} & {} & {} \\
\midrule

Aircraft & 49.2 $ \pm $ 0.1 & 50.0 $ \pm $ 0.1 & \textbf{55.1 $ \pm $ 0.1} & 52.5 $ \pm $ 0.2 & {N \slash A} & 54.6 $ \pm $ 0.1 & \textbf{55.2 $ \pm $ 0.2} \\
Birds & 57.7 $ \pm $ 0.1 & 57.1 $ \pm $ 0.0 & 58.4 $ \pm $ 0.0 & \textbf{59.2 $ \pm $ 0.0} & {N \slash A} & 55.2 $ \pm $ 0.0 & 56.7 $ \pm $ 0.1 \\
CIFAR-10 & 91.2 $ \pm $ 0.0 & 90.2 $ \pm $ 0.0 & 90.9 $ \pm $ 0.0 & \textbf{91.4 $ \pm $ 0.0} & {N \slash A} & 89.7 $ \pm $ 0.1 & 90.0 $ \pm $ 0.1 \\
CIFAR-100 & 74.6 $ \pm $ 0.1 & 73.7 $ \pm $ 0.0 & 74.7 $ \pm $ 0.1 & \textbf{75.3 $ \pm $ 0.0} & {N \slash A} & 73.1 $ \pm $ 0.1 & 73.7 $ \pm $ 0.0 \\
Caltech-101 & 91.9 $ \pm $ 0.1 & \textbf{92.1 $ \pm $ 0.2} & \textbf{92.4 $ \pm $ 0.2} & \textbf{92.3 $ \pm $ 0.2} & {N \slash A} & 91.1 $ \pm $ 0.1 & 90.8 $ \pm $ 0.3 \\
Caltech-256 & 84.8 $ \pm $ 0.1 & 84.6 $ \pm $ 0.1 & 84.6 $ \pm $ 0.1 & \textbf{85.4 $ \pm $ 0.1} & {N \slash A} & 83.3 $ \pm $ 0.1 & 84.6 $ \pm $ 0.1 \\
Cars & 53.4 $ \pm $ 0.1 & 51.4 $ \pm $ 0.1 & 56.6 $ \pm $ 0.0 & 56.2 $ \pm $ 0.1 & {N \slash A} & 57.4 $ \pm $ 0.1 & \textbf{58.6 $ \pm $ 0.1} \\
DTD & 73.5 $ \pm $ 0.2 & 73.1 $ \pm $ 0.2 & \textbf{74.4 $ \pm $ 0.1} & \textbf{74.1 $ \pm $ 0.2} & {N \slash A} & 73.5 $ \pm $ 0.2 & 72.9 $ \pm $ 0.3 \\
Flowers & 91.6 $ \pm $ 0.1 & 91.1 $ \pm $ 0.1 & \textbf{92.7 $ \pm $ 0.1} & \textbf{92.6 $ \pm $ 0.1} & {N \slash A} & 92.2 $ \pm $ 0.1 & 92.3 $ \pm $ 0.1 \\
Food-101 & 73.2 $ \pm $ 0.0 & 72.2 $ \pm $ 0.0 & 73.8 $ \pm $ 0.0 & \textbf{74.8 $ \pm $ 0.0} & {N \slash A} & 73.3 $ \pm $ 0.0 & 72.5 $ \pm $ 0.1 \\
Pets & \textbf{92.6 $ \pm $ 0.1} & 91.9 $ \pm $ 0.1 & 92.3 $ \pm $ 0.1 & \textbf{92.8 $ \pm $ 0.1} & {N \slash A} & 91.9 $ \pm $ 0.1 & \textbf{92.5 $ \pm $ 0.2} \\
SUN397 & 60.1 $ \pm $ 0.0 & 59.9 $ \pm $ 0.0 & 60.4 $ \pm $ 0.0 & \textbf{60.5 $ \pm $ 0.0} & {N \slash A} & 59.1 $ \pm $ 0.1 & 59.9 $ \pm $ 0.0 \\

\midrule
{90\% Sparsity} & {} & {} & {} & {} & {} & {} & {} \\
\midrule

Aircraft & 49.2 $ \pm $ 0.1 & 50.0 $ \pm $ 0.1 & 55.5 $ \pm $ 0.1 & 54.6 $ \pm $ 0.1 & 55.5 $ \pm $ 0.0 & 54.1 $ \pm $ 0.1 & \textbf{56.6 $ \pm $ 0.1} \\
Birds & 57.7 $ \pm $ 0.1 & 57.1 $ \pm $ 0.0 & 58.7 $ \pm $ 0.0 & 59.7 $ \pm $ 0.1 & \textbf{60.4 $ \pm $ 0.1} & 53.3 $ \pm $ 0.0 & 57.2 $ \pm $ 0.1 \\
CIFAR-10 & \textbf{91.2 $ \pm $ 0.0} & 90.2 $ \pm $ 0.0 & 91.0 $ \pm $ 0.0 & 90.9 $ \pm $ 0.0 & 90.3 $ \pm $ 0.0 & 90.0 $ \pm $ 0.1 & 90.2 $ \pm $ 0.1 \\
CIFAR-100 & \textbf{74.6 $ \pm $ 0.1} & 73.7 $ \pm $ 0.0 & 74.3 $ \pm $ 0.0 & \textbf{74.7 $ \pm $ 0.0} & 74.2 $ \pm $ 0.0 & 72.8 $ \pm $ 0.1 & 73.4 $ \pm $ 0.1 \\
Caltech-101 & 91.9 $ \pm $ 0.1 & 92.1 $ \pm $ 0.2 & 92.5 $ \pm $ 0.1 & \textbf{92.9 $ \pm $ 0.2} & \textbf{92.8 $ \pm $ 0.2 } & 90.6 $ \pm $ 0.3 & 91.4 $ \pm $ 0.4 \\
Caltech-256 & 84.8 $ \pm $ 0.1 & 84.6 $ \pm $ 0.1 & 84.5 $ \pm $ 0.0 & \textbf{85.3 $ \pm $ 0.1} & \textbf{85.3 $ \pm $ 0.1} & 81.9 $ \pm $ 0.0 & 84.5 $ \pm $ 0.1 \\
Cars & 53.4 $ \pm $ 0.1 & 51.4 $ \pm $ 0.1 & 56.0 $ \pm $ 0.1 & 58.7 $ \pm $ 0.1 & 58.4 $ \pm $ 0.1 & 55.5 $ \pm $ 0.1 & \textbf{60.5 $ \pm $ 0.1} \\
DTD & 73.5 $ \pm $ 0.2 & 73.1 $ \pm $ 0.2 & 73.7 $ \pm $ 0.2 & \textbf{74.2 $ \pm $ 0.3} & \textbf{73.9 $ \pm $ 0.2} & 72.6 $ \pm $ 0.3 & 72.7 $ \pm $ 0.2 \\
Flowers & 91.6 $ \pm $ 0.1 & 91.1 $ \pm $ 0.1 & 92.4 $ \pm $ 0.0 & 92.6 $ \pm $ 0.0 & \textbf{92.8 $ \pm $ 0.0} & 91.6 $ \pm $ 0.1 & 92.4 $ \pm $ 0.1 \\
Food-101 & 73.2 $ \pm $ 0.0 & 72.2 $ \pm $ 0.0 & 73.8 $ \pm $ 0.0 & 75.1 $ \pm $ 0.0 & \textbf{75.2 $ \pm $ 0.0} & 71.7 $ \pm $ 0.0 & 72.6 $ \pm $ 0.0 \\
Pets & 92.6 $ \pm $ 0.1 & 91.9 $ \pm $ 0.1 & 91.9 $ \pm $ 0.1 & \textbf{92.5 $ \pm $ 0.1} & \textbf{92.6 $ \pm $ 0.2} & 91.1 $ \pm $ 0.1 & 91.9 $ \pm $ 0.2 \\
SUN397 & 60.1 $ \pm $ 0.0 & 59.9 $ \pm $ 0.0 & 59.8 $ \pm $ 0.1 & 60.3 $ \pm $ 0.0 & \textbf{61.2 $ \pm $ 0.0} & 57.7 $ \pm $ 0.0 & 59.8 $ \pm $ 0.1 \\

\midrule
{95\% Sparsity} & {} & {} & {} & {} & {} & {} & {} \\
\midrule
Aircraft & 49.2 $ \pm $ 0.1 & 50.0 $ \pm $ 0.1 & 56.6 $ \pm $ 0.1 & 55.6 $ \pm $ 0.0 & {N \slash A} & 53.5 $ \pm $ 0.1 & \textbf{56.9 $ \pm $ 0.1} \\
Birds & 57.7 $ \pm $ 0.1 & 57.1 $ \pm $ 0.0 & 57.7 $ \pm $ 0.0 & \textbf{59.2 $ \pm $ 0.1} & {N \slash A} & 51.9 $ \pm $ 0.1 & 55.9 $ \pm $ 0.0 \\
CIFAR-10 & \textbf{91.2 $ \pm $ 0.0} & 90.2 $ \pm $ 0.0 & 90.5 $ \pm $ 0.0 & 90.2 $ \pm $ 0.0 & {N \slash A} & 89.4 $ \pm $ 0.0 & 89.8 $ \pm $ 0.1 \\
CIFAR-100 & \textbf{74.6 $ \pm $ 0.1} & 73.7 $ \pm $ 0.0 & 73.4 $ \pm $ 0.0 & 74.3 $ \pm $ 0.1 & {N \slash A} & 71.5 $ \pm $ 0.1 & 72.4 $ \pm $ 0.1 \\
Caltech-101 & \textbf{91.9 $ \pm $ 0.1} & \textbf{92.1 $ \pm $ 0.2} & 91.6 $ \pm $ 0.1 & \textbf{92.3 $ \pm $ 0.3} & {N \slash A} & 89.0 $ \pm $ 0.1 & 91.4 $ \pm $ 0.1 \\
Caltech-256 & \textbf{84.8 $ \pm $ 0.1} & \textbf{84.6 $ \pm $ 0.1} & 82.8 $ \pm $ 0.1 & 84.0 $ \pm $ 0.1 & {N \slash A} & 80.1 $ \pm $ 0.1 & 83.5 $ \pm $ 0.1 \\
Cars & 53.4 $ \pm $ 0.1 & 51.4 $ \pm $ 0.1 & 56.9 $ \pm $ 0.1 & \textbf{57.4 $ \pm $ 0.0} & {N \slash A} & 52.9 $ \pm $ 0.0 & 57.0 $ \pm $ 0.1 \\
DTD & 73.5 $ \pm $ 0.2 & 73.1 $ \pm $ 0.2 & 72.7 $ \pm $ 0.1 & \textbf{74.7 $ \pm $ 0.2} & {N \slash A} & 71.9 $ \pm $ 0.1 & 72.9 $ \pm $ 0.2 \\
Flowers & 91.6 $ \pm $ 0.1 & 91.1 $ \pm $ 0.1 & \textbf{93.0 $ \pm $ 0.1} & 92.5 $ \pm $ 0.1 & {N \slash A} & 91.0 $ \pm $ 0.1 & 92.4 $ \pm $ 0.1 \\
Food-101 & 73.2 $ \pm $ 0.0 & 72.2 $ \pm $ 0.0 & 73.2 $ \pm $ 0.0 & \textbf{74.7 $ \pm $ 0.0} & {N \slash A} & 70.6 $ \pm $ 0.1 & 71.9 $ \pm $ 0.0 \\
Pets & \textbf{92.6 $ \pm $ 0.1} & 91.9 $ \pm $ 0.1 & 91.0 $ \pm $ 0.2 & 91.5 $ \pm $ 0.1 & {N \slash A} & 90.1 $ \pm $ 0.1 & 91.1 $ \pm $ 0.1 \\
SUN397 & \textbf{60.1 $ \pm $ 0.0} & 59.9 $ \pm $ 0.0 & 58.2 $ \pm $ 0.0 & 59.7 $ \pm $ 0.0 & {N \slash A} & 55.9 $ \pm $ 0.1 & 58.3 $ \pm $ 0.1 \\
\bottomrule
\end{tabular}

}
\caption{Transfer accuracy for \emph{extended training time} for ResNet50 with \emph{linear finetuning}.}
\label{table:rn50_linear_longruns}
\end{table}

%% file: tables/rn50_full_longruns.tex
\begin{table}[htb!]
\centering
\scalebox{0.9}{
\begin{tabular}{ccc|ccc|cc}
\toprule
{} & {Dense} & {Dense 2x} & {AC/DC} & {AC/DC 3x} & {AC/DC 5x} & {RigL ERK 1x} & {RigL ERK 5x} \\
\midrule
{80\% Sparsity} & {} & {} & {} & {} & {} & {} & {} \\
\midrule

Aircraft & 83.6 $ \pm $ 0.4 & \textbf{84.3 $ \pm $ 0.2} & 83.3 $ \pm $ 0.1 & 83.2 $ \pm $ 0.3 & {N \slash A} & 82.6 $ \pm $ 0.3 & 82.4 $ \pm $ 0.2 \\
Birds & 72.4 $ \pm $ 0.3 & \textbf{73.5 $ \pm $ 0.1} & 69.9 $ \pm $ 0.2 & 72.5 $ \pm $ 0.2 & {N \slash A} & 72.3 $ \pm $ 0.3 & \textbf{73.4 $ \pm $ 0.1} \\
CIFAR-10 & \textbf{97.4 $ \pm $ 0.0} & \textbf{97.4 $ \pm $ 0.0} & 96.9 $ \pm $ 0.1 & \textbf{97.5 $ \pm $ 0.1} & {N \slash A} & 96.9 $ \pm $ 0.0 & 97.1 $ \pm $ 0.0 \\
CIFAR-100 & \textbf{85.6 $ \pm $ 0.2} & \textbf{85.8 $ \pm $ 0.2} & 84.9 $ \pm $ 0.2 & 85.3 $ \pm $ 0.1 & {N \slash A} & 83.6 $ \pm $ 0.2 & 84.1 $ \pm $ 0.4 \\
Caltech-101 & 93.5 $ \pm $ 0.1 & \textbf{93.9 $ \pm $ 0.1} & 92.5 $ \pm $ 0.2 & 93.4 $ \pm $ 0.2 & {N \slash A} & 92.5 $ \pm $ 0.1 & 92.0 $ \pm $ 0.3 \\
Caltech-256 & 86.1 $ \pm $ 0.1 & \textbf{86.5 $ \pm $ 0.2} & 85.4 $ \pm $ 0.2 & \textbf{86.7 $ \pm $ 0.1} & {N \slash A} & 83.8 $ \pm $ 0.1 & 84.2 $ \pm $ 0.2 \\
Cars & \textbf{90.3 $ \pm $ 0.2} & \textbf{90.5 $ \pm $ 0.2} & 89.2 $ \pm $ 0.1 & 89.8 $ \pm $ 0.3 & {N \slash A} & 89.4 $ \pm $ 0.1 & 89.6 $ \pm $ 0.1 \\
DTD & 76.2 $ \pm $ 0.3 & \textbf{76.9 $ \pm $ 0.3} & 75.7 $ \pm $ 0.5 & 76.6 $ \pm $ 0.0 & {N \slash A} & 74.5 $ \pm $ 0.2 & 74.2 $ \pm $ 0.2 \\
Flowers & 95.0 $ \pm $ 0.1 & 95.5 $ \pm $ 0.2 & 94.7 $ \pm $ 0.2 & 95.1 $ \pm $ 0.2 & {N \slash A} & 95.7 $ \pm $ 0.2 & \textbf{96.1 $ \pm $ 0.1} \\
Food-101 & 87.3 $ \pm $ 0.1 & \textbf{87.5 $ \pm $ 0.1} & 86.9 $ \pm $ 0.1 & \textbf{87.7 $ \pm $ 0.1} & {N \slash A} & 86.9 $ \pm $ 0.1 & \textbf{87.2 $ \pm $ 0.1} \\
Pets & \textbf{93.4 $ \pm $ 0.1} & \textbf{93.4 $ \pm $ 0.2} & 92.5 $ \pm $ 0.0 & \textbf{93.4 $ \pm $ 0.2} & {N \slash A} & 92.2 $ \pm $ 0.1 & 92.4 $ \pm $ 0.1 \\
SUN397 & 64.8 $ \pm $ 0.0 & \textbf{65.1 $ \pm $ 0.0} & 64.0 $ \pm $ 0.0 & 64.8 $ \pm $ 0.1 & {N \slash A} & 62.2 $ \pm $ 0.2 & 62.0 $ \pm $ 0.3 \\

\midrule
{90\% Sparsity} & {} & {} & {} & {} & {} & {} & {} \\
\midrule

Aircraft & 83.6 $ \pm $ 0.4 & \textbf{84.3 $ \pm $ 0.2} & 82.8 $ \pm $ 1.0 & 82.6 $ \pm $ 0.2 & \textbf{83.5 $ \pm $ 0.4} & 81.6 $ \pm $ 0.5 & 83.0 $ \pm $ 0.4 \\
Birds & 72.4 $ \pm $ 0.3 & \textbf{73.5 $ \pm $ 0.1} & 68.5 $ \pm $ 0.1 & 71.6 $ \pm $ 0.2 & 72.8 $ \pm $ 0.2 & 70.3 $ \pm $ 0.0 & 72.9 $ \pm $ 0.2 \\
CIFAR-10 & \textbf{97.4 $ \pm $ 0.0} & \textbf{97.4 $ \pm $ 0.0} & 96.6 $ \pm $ 0.1 & 97.0 $ \pm $ 0.1 & 97.1 $ \pm $ 0.1 & 96.4 $ \pm $ 0.1 & 97.0 $ \pm $ 0.1 \\
CIFAR-100 & \textbf{85.6 $ \pm $ 0.2} & \textbf{85.8 $ \pm $ 0.2} & 83.9 $ \pm $ 0.1 & 84.7 $ \pm $ 0.1 & 85.3 $ \pm $ 0.1 & 83.0 $ \pm $ 0.2 & 83.7 $ \pm $ 0.3 \\
Caltech-101 & 93.5 $ \pm $ 0.1 & \textbf{93.9 $ \pm $ 0.1} & 92.6 $ \pm $ 0.2 & 93.0 $ \pm $ 0.0 & 93.2 $ \pm $ 0.1 & 91.7 $ \pm $ 0.3 & 92.3 $ \pm $ 0.4 \\
Caltech-256 & 86.1 $ \pm $ 0.1 & \textbf{86.5 $ \pm $ 0.2} & 84.8 $ \pm $ 0.1 & 86.1 $ \pm $ 0.1 & \textbf{86.5 $ \pm $ 0.1} & 82.7 $ \pm $ 0.2 & 84.0 $ \pm $ 0.1 \\
Cars & \textbf{90.3 $ \pm $ 0.2} & \textbf{90.5 $ \pm $ 0.2} & 88.5 $ \pm $ 0.2 & 89.3 $ \pm $ 0.1 & 89.8 $ \pm $ 0.1 & 88.4 $ \pm $ 0.1 & 89.2 $ \pm $ 0.1 \\
DTD & 76.2 $ \pm $ 0.3 & \textbf{76.9 $ \pm $ 0.3} & 75.2 $ \pm $ 0.1 & 75.3 $ \pm $ 0.2 & 76.4 $ \pm $ 0.2 & 73.4 $ \pm $ 0.4 & 75.2 $ \pm $ 0.8 \\
Flowers & 95.0 $ \pm $ 0.1 & 95.5 $ \pm $ 0.2 & 94.6 $ \pm $ 0.1 & 95.4 $ \pm $ 0.1 & \textbf{95.9 $ \pm $ 0.1} & 95.5 $ \pm $ 0.1 & \textbf{96.1 $ \pm $ 0.1} \\
Food-101 & 87.3 $ \pm $ 0.1 & \textbf{87.5 $ \pm $ 0.1} & 86.6 $ \pm $ 0.1 & 87.4 $ \pm $ 0.1 & \textbf{87.7 $ \pm $ 0.1} & 85.9 $ \pm $ 0.1 & 87.3 $ \pm $ 0.2 \\
Pets & \textbf{93.4 $ \pm $ 0.1} & \textbf{93.4 $ \pm $ 0.2} & 92.1 $ \pm $ 0.1 & 92.6 $ \pm $ 0.1 & 93.0 $ \pm $ 0.1 & 91.4 $ \pm $ 0.2 & 92.3 $ \pm $ 0.1 \\
SUN397 & 64.8 $ \pm $ 0.0 & \textbf{65.1 $ \pm $ 0.0} & 63.0 $ \pm $ 0.0 & 64.3 $ \pm $ 0.0 & 64.8 $ \pm $ 0.1 & 61.3 $ \pm $ 0.1 & 62.0 $ \pm $ 0.2 \\

\midrule
{95\% Sparsity} & {} & {} & {} & {} & {} & {} & {} \\
\midrule
Aircraft & 83.6 $ \pm $ 0.4 & \textbf{84.3 $ \pm $ 0.2} & 81.2 $ \pm $ 0.4 & 82.2 $ \pm $ 0.3 & {N \slash A} & 80.7 $ \pm $ 0.1 & 82.5 $ \pm $ 0.4 \\
Birds & 72.4 $ \pm $ 0.3 & \textbf{73.5 $ \pm $ 0.1} & 66.9 $ \pm $ 0.1 & 70.1 $ \pm $ 0.1 & {N \slash A} & 68.3 $ \pm $ 0.2 & 71.6 $ \pm $ 0.1 \\
CIFAR-10 & \textbf{97.4 $ \pm $ 0.0} & \textbf{97.4 $ \pm $ 0.0} & 96.2 $ \pm $ 0.1 & 96.6 $ \pm $ 0.1 & {N \slash A} &96.0 $ \pm $ 0.1 & 96.6 $ \pm $ 0.1 \\
CIFAR-100 & \textbf{85.6 $ \pm $ 0.2} & \textbf{85.8 $ \pm $ 0.2} & 82.9 $ \pm $ 0.1 & 84.0 $ \pm $ 0.1 & {N \slash A} &82.0 $ \pm $ 0.2 & 82.8 $ \pm $ 0.0 \\
Caltech-101 & 93.5 $ \pm $ 0.1 & \textbf{93.9 $ \pm $ 0.1} & 91.9 $ \pm $ 0.2 & 92.9 $ \pm $ 0.1 & {N \slash A} &90.7 $ \pm $ 0.4 & 92.2 $ \pm $ 0.3 \\
Caltech-256 & 86.1 $ \pm $ 0.1 & \textbf{86.5 $ \pm $ 0.2} & 83.1 $ \pm $ 0.0 & 85.3 $ \pm $ 0.0 & {N \slash A} &81.1 $ \pm $ 0.2 & 83.1 $ \pm $ 0.2 \\
Cars & \textbf{90.3 $ \pm $ 0.2} & \textbf{90.5 $ \pm $ 0.2} & 87.6 $ \pm $ 0.1 & 89.0 $ \pm $ 0.1 & {N \slash A} &87.9 $ \pm $ 0.3 & 88.9 $ \pm $ 0.2 \\
DTD & 76.2 $ \pm $ 0.3 & \textbf{76.9 $ \pm $ 0.3} & 74.1 $ \pm $ 0.4 & 75.0 $ \pm $ 0.4 & {N \slash A} &73.3 $ \pm $ 0.2 & 73.5 $ \pm $ 0.2 \\
Flowers & 95.0 $ \pm $ 0.1 & 95.5 $ \pm $ 0.2 & 94.1 $ \pm $ 0.3 & 95.0 $ \pm $ 0.2 & {N \slash A} &94.9 $ \pm $ 0.3 & \textbf{96.0 $ \pm $ 0.0} \\
Food-101 & \textbf{87.3 $ \pm $ 0.1} & \textbf{87.5 $ \pm $ 0.1} & 85.5 $ \pm $ 0.0 & 86.7 $ \pm $ 0.1 & {N \slash A} &85.1 $ \pm $ 0.2 & 86.6 $ \pm $ 0.0 \\
Pets & \textbf{93.4 $ \pm $ 0.1} & \textbf{93.4 $ \pm $ 0.2} & 91.0 $ \pm $ 0.1 & 91.6 $ \pm $ 0.1 & {N \slash A} &90.1 $ \pm $ 0.2 & 91.6 $ \pm $ 0.3 \\
SUN397 & 64.8 $ \pm $ 0.0 & \textbf{65.1 $ \pm $ 0.0} & 61.4 $ \pm $ 0.2 & 63.0 $ \pm $ 0.0 & {N \slash A} &60.0 $ \pm $ 0.3 & 61.1 $ \pm $ 0.2 \\
\bottomrule
\end{tabular}

}
\caption{Transfer accuracy for \emph{extended training time} for ResNet50 with \emph{full finetuning}.}
\label{table:rn50_full_longruns}
\end{table}